\documentclass{article}

\usepackage[main, final]{neurips_2026}
\usepackage[utf8]{inputenc} 
\usepackage[T1]{fontenc}    
\usepackage[hidelinks]{hyperref}       
\usepackage{url}            
\usepackage{booktabs}       
\usepackage{multirow}       
\usepackage{longtable}      
\usepackage{amsmath}        
\usepackage{amsfonts}       
\usepackage{nicefrac}       
\usepackage{standalone}

\usepackage{microtype}      
\usepackage{xcolor}         
\usepackage{graphicx}       
\usepackage{wrapfig}        
\usepackage{subcaption}     
\usepackage{algorithm}      
\usepackage{algorithmic}    
\newcommand{\algstep}[1]{\item[]\textbf{#1}}

\usepackage{tikz}
\usetikzlibrary{arrows.meta,positioning,calc,decorations.pathreplacing}
\usetikzlibrary{arrows.meta,calc,positioning}

\usepackage{graphicx}
\usepackage{amsmath,amssymb}
\usepackage[capitalize,nameinlink]{cleveref} 

\DeclareRobustCommand{\MANCE}{MANCE}
\DeclareRobustCommand{\MANCEp}{MANCE\textsuperscript{+}}
\DeclareRobustCommand{\MANCEpp}{MANCE\textsuperscript{++}}

\DeclareRobustCommand{\AmbCEpp}{AmbCE\textsuperscript{++}}

\makeatother

\title{MANCE: Manifold Aware Concept Erasure}

\author{
  Matan Avitan \\
  Bar-Ilan University \\
  \texttt{matan13av@gmail.com} \\
  \And
  Yoav Goldberg \\
  Bar-Ilan University \\
  Allen Institute for Artificial Intelligence \\
  \texttt{yoav.goldberg@gmail.com} \\
  \And
  Yanai Elazar \\
  Bar-Ilan University \\
  \texttt{yanaiela@gmail.com} \\
}

\begin{document}

\maketitle

\section*{Abstract}

Concept erasure aims to remove a target concept from a representation while preserving the other information encoded in it. 
This is difficult because representations encode many concepts that are often correlated with the erasure target, so removing the target risks damaging them.
We propose the \emph{Manifold Constraint Hypothesis} (MCH): if natural representations concentrate on a structured, lower-dimensional manifold, then interventions should be constrained to that manifold and better preserve other information encoded in the representation during interventions.
We instantiate MCH in a new concept erasure method: MANifold aware Concept Erasure (\emph{\MANCE{}}). \MANCE{} performs iterative updates to the representations using signals from a classifier that predicts a target concept.
We estimate the manifold using representations obtained from natural inputs, and then we project the concept removal update to the estimated manifold.
We perform extensive evaluation on 119 settings spanning text and vision, including 13 language models, three NLP concepts, and 40 CelebA-CLIP attributes. Employing \MANCE{} on top of previous methods shows consistent improved leakage results.
We also introduce \MANCEp{} and \MANCEpp{}, which prepend a closed-form erasure algorithm before employing \MANCE{}, achieving better \textit{leakage}--\textit{surgicality} tradeoffs relative to matched full-space updates.
\MANCEpp{}, our best method, achieves state-of-the-art results on nonlinear concept erasure. These results support MCH in the erasure setting: interventions should be constrained to the natural representation manifold.\footnote{Code is available at: \url{https://github.com/MatanAvitan/mance}}

\section{Introduction}
\label{sec:intro}

Neural representations encode many attributes of their inputs: semantic content, style, demographic information, task labels, and other latent factors \citep{cram_vectors_conneau,nlp_bert-pipeline_tenney,syntax_goldberg2019,structural-probe,diagnostic_adi,rogers-etal-2020-primer,embeddings_bias,elazar-etal-2021-amnesic}. In many applications, we want to remove one such attribute (e.g., gender or toxicity), while preserving everything else the representation contains \citep{dearteaga-etal-2019-bias,avitan-etal-2025-practical}. This is the central tension in \textit{concept erasure}: removing information about the attribute selected for erasure without unnecessarily damaging other concepts, known as \textit{surgicality} \citep{singh-ravfogel-2024-representation-surgery}.

Intervening on representations is difficult. Since concepts are often entangled with one another (e.g., profession information and gender), due to data correlation and \emph{superposition} \citep{elhage2022toy}, intervening on one concept may negatively affect another.

A further challenge is that these other concepts are usually unknown.
This makes it hard to formulate erasure as a standard constrained optimization problem to remove a concept while preserving others. 

We propose the \emph{Manifold Constraint Hypothesis} (MCH), formalized in \Cref{sec:mch}. 
MCH assumes that representations that arrive from the natural input distribution lie on a manifold. 
MCH then predicts that interventions constrained to that manifold will be equally effective in concept erasure, while also preserving other concepts encoded in the representation.

\begin{figure*}
  \centering
  \resizebox{\linewidth}{!}{\includegraphics{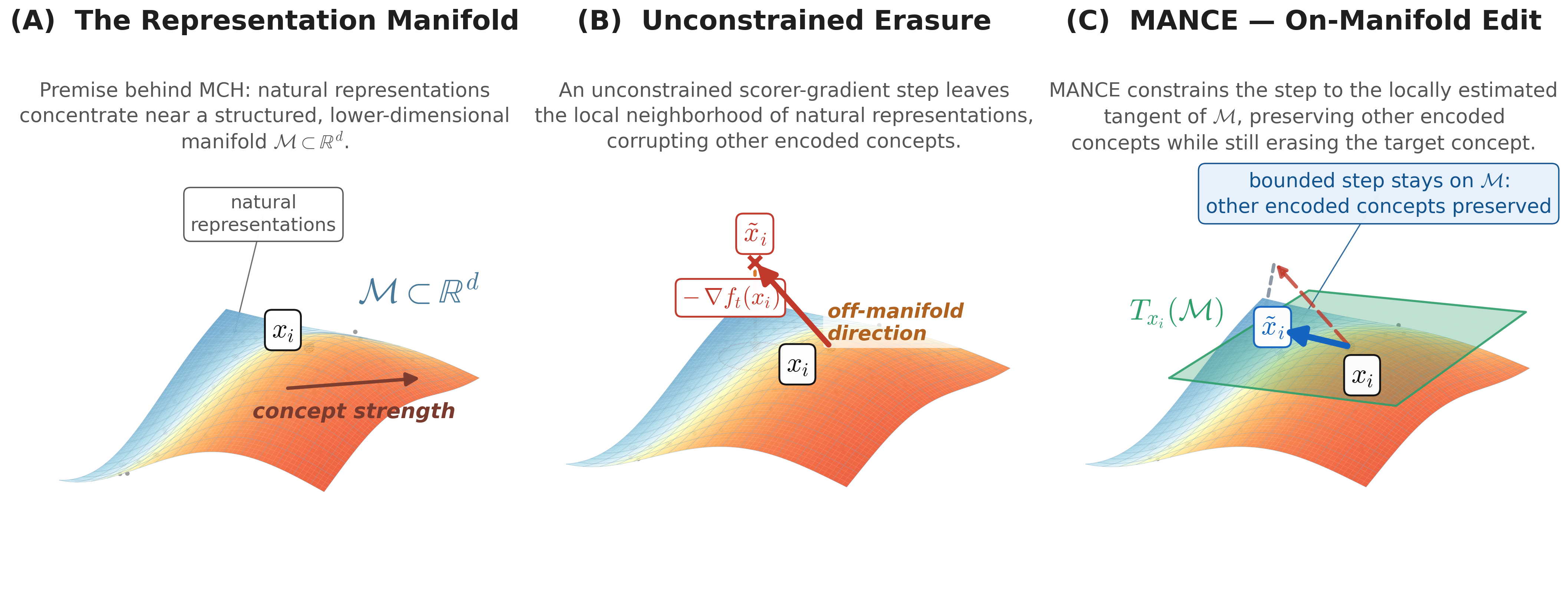}}
  \caption{\textbf{MCH motivates erasure constrained by the representation manifold.} The \emph{Manifold Constraint Hypothesis} assumes that natural representations concentrate in a structured, lower-dimensional manifold. Under this premise, an unconstrained intervention can change an encoded concept while moving a representation away from the local neighborhood of natural representations, potentially damaging unlabeled control information encoded along the manifold. \MANCE{} operationalizes MCH for concept erasure by estimating the manifold, and projecting the erasure direction onto the manifold. 
}
  \label{fig:mance}
\end{figure*}

We introduce the \MANCE{} family, a series of manifold aware methods for non-linear concept erasure \citep{edwards2016censoring, elazar-goldberg-2018-adversarial, ravfogel-etal-2020-null, ravfogel2022linear, ravfogel-etal-2022-adversarial} that operationalizes MCH. 
The main idea is to perform a targeted erasure by modifying the representations based on some target concept we would like to erase, and constrain the update to remain on the manifold.
We implement the targeted erasure using a standard gradient-based approach (e.g., \cite{Goodfellow2015ExplainingAH}). Then, to perform the constrained update we use local manifold estimation by its neighboring representations obtained from natural inputs \citep{zhang-zha-2004-ltsa}.
As part of the constrained optimization, we develop a closed-form solution that allows a flexible update based on the manifold's local geometry.
We show that \MANCE{} can be effectively added to any previous erasure method, by applying it on the intervened representations.
We also propose two variants of \MANCE{} by prepending closed-form linear stages: \MANCEp{} prepends LEACE \citep{belrose-etal-2023-leace}; and \MANCEpp{} further prepends a rank-2 projection that removes the leading second-moment covariance asymmetry.

We perform a comprehensive evaluation of concept erasure: 119 settings across text and vision, including 39 NLP settings spanning 13 language models and three concepts, plus 80 CelebA-CLIP settings from 40 image attributes under two surgicality regimes. 
By applying \MANCE{} on previous erasure methods (INLP \citep{ravfogel-etal-2020-null}, LEACE \citep{belrose-etal-2023-leace}, and IGBP \citep{iskander-etal-2023-shielded}), we show consistent erasure improvements while minimally harming other concepts.
Finally, we show that \MANCEpp{} achieves state-of-the-art results on concept erasure, outperforming the strongest baseline \citep{akbari-etal-2025-obliviator} under the same limits on control accuracy degradation.

We make the following contributions:
(1) We formalize MCH, which predicts that interventions constrained to the natural representations manifold preserve other concepts encoded in the representation while still effectively intervening on the intended concept; (2) we operationalize this prediction for concept erasure through \MANCE{}, a gradient-based erasure method w.r.t. the natural representation manifold, and show that it improves prior erasers performance when applied as an additional step; and (3) we introduce \MANCEpp{}, a composed variant that adds closed-form preprocessing before \MANCE{} and achieves state-of-the-art results on concept-erasure.

\section{Previous Work}
\label{sec:related}

Concept erasure methods aim to remove information about a target concept from neural representations while minimizing damage to the other information those representations encode.
Linear methods such as INLP \citep{ravfogel-etal-2020-null} and LEACE \citep{belrose-etal-2023-leace} target linearly decodable concept information. 
IGBP \citep{iskander-etal-2023-shielded} and Obliviator \citep{akbari-etal-2025-obliviator} remove nonlinear information.
IGBP iteratively trains a nonlinear probe and retracts representations toward the current decision boundary; Obliviator approximates a kernel feature map with random Fourier features and iteratively projects out the directions in that feature space that are most predictive of the target concept.
A practical limitation of Obliviator is that its erased output is not guaranteed to preserve the original representation dimensionality: the method selects retained feature-space modes through an eigenvalue problem, and the resulting dimension depends on the retained modes. This does not prevent probe-based evaluation, but it limits as-is use in downstream settings where the edited vector must have the original activation dimension before it can be inserted back into the model computation, such as activation patching.

\MANCE{} is orthogonal to the advances in concept erasure in the sense that it can be applied to any previous erasure method.
Rather than move in arbitrary directions, \MANCE{} projects the erasure update onto the natural-representation manifold. As such, we employ \MANCE{} in addition to previous concept erasure methods, and demonstrate over a wide range of settings how the manifold constraint improves the surgicality of the edits.

\section{The \MANCE{} Algorithms Family}
\label{sec:method}

\subsection{The Manifold Constraint Hypothesis}
\label{sec:mch}

Hidden states produced by natural inputs do not fill the full representation space uniformly. We assume that they instead occupy a structured, lower-dimensional region shaped by the encoder and the data distribution, which we model as a manifold $\mathcal{M} \subset \mathbb{R}^d$, where $d$ is the representation dimension.

We call an intervention $\mathbf{x} \mapsto \mathbf{x} + \boldsymbol{\delta}$ \emph{constrained to the manifold} when it moves the representation along $\mathcal{M}$, so that the edited point $\mathbf{x} + \boldsymbol{\delta}$ remains on $\mathcal{M}$. This condition is stated with respect to $\mathcal{M}$ itself, independent of how the manifold is modeled or estimated. By contrast, an intervention is \emph{unconstrained} when $\boldsymbol{\delta}$ may point anywhere in $\mathbb{R}^d$, at any magnitude.

The \emph{Manifold Constraint Hypothesis} (MCH) posits that, among interventions with a matched effect on the target concept, manifold-constrained interventions preserve other concepts encoded in the representation better than unconstrained interventions.
\MANCE{}, presented next, operationalizes MCH by estimating a first-order approximation to the manifold using tangent directions obtained from natural inputs.

\subsection{The \MANCE{} method}

\emph{MANifold aware Concept Erasure} (\MANCE{}) operationalizes MCH (\Cref{sec:mch}) as a manifold-constrained erasure operation on representations. Given initial representations $\mathbf{X}^{(0)} \in \mathbb{R}^{N \times d}$, where $N$ is the number of samples, $d$ is the representation dimension, row $i$ is $\mathbf{x}_i^{(0)}$, and the target-concept labels are $\mathbf{y}=(y_1,\ldots,y_N)$, \MANCE{} fits a nonlinear concept probe $f_t : \mathbb{R}^d \to \mathbb{R}$ at each round $t$, computes the gradient of the probe's target prediction with respect to each input representation, and edits only the locally manifold-supported part of that gradient. 
We present the method for binary target labels.\footnote{Multiclass targets can be handled by fitting one-vs-rest probes for the class to erase.}  
The method has three ingredients: (1) estimate the local manifold from representations $\mathbf{X}^{(0)}$ obtained from natural inputs; (2) project the erasure gradient onto the span of those local directions; and (3) choose a per-sample step size, small enough that the edit stays inside the manifold.

\paragraph{Notation.} Representations $\mathbf{x}_i$ lie in $\mathbb{R}^d$. We write $\mathcal{M}$ for the representation manifold and $T_{\mathbf{x}_i}(\mathcal{M})$ for its \emph{tangent space} at $\mathbf{x}_i$: the linear subspace that approximates $\mathcal{M}$ to first order near $\mathbf{x}_i$, i.e.,\ the directions one can move along while remaining on $\mathcal{M}$ \citep{tu-2011-manifolds}. The method is controlled by the following scalars: $H$, the number of erasure rounds; $\tau$, the probe-refit period (defined below); $k$, the neighborhood size; $r$, the number of local directions retained (the tangent rank); $\lambda_i$, the per-sample step size; $\varepsilon$, the update size as a fraction of the local neighborhood radius; $\lambda_{\max}$, a cap on any single step; and $\alpha$, the spectral-weighting exponent.
A \emph{probe refit} means retraining the MLP probe from scratch on the current edited representations and the same target labels. We do this every $\tau$ rounds, rather than every round, for efficiency; we fix $\tau=8$ across all settings (see App.~\ref{sec:appendix_hyperparameters}). Refitting matters because after an edit removes the currently easiest target-concept signal, a new nonlinear probe can expose residual information about the erased concept that the next tangent-constrained step should remove.

\paragraph{Summary of \MANCE{}.}

\MANCE{} is an iterative algorithm for nonlinear concept erasure. At each step it computes a nonlinear concept probe and obtains its gradients w.r.t. the inputs. Crucially, it constrains the gradient update to the manifold of the natural representation (produced by natural inputs, not related to the representations we remove the concept from). Concretely, it repeatedly estimates a local tangent basis (using natural inputs representation) around each current representation to edit, projects the concept probe's gradient into that basis, reweights the tangent-basis coordinates by the local singular spectrum, and applies the largest per-sample update allowed by the local neighborhood radius. At a high level, each round uses the current nonlinear concept probe, recomputes local tangent bases, applies the per-sample update to every representation, and returns $\mathbf{X}^{(H)}$ after $H$ rounds. 

\paragraph{Step 1: Estimate the manifold locally.}

The first step is to estimate the representation manifold that will constrain the edits. The representation manifold is unknown, but estimating it globally is unnecessary for an edit at one point. For each representation $\mathbf{x}_i$, we instead estimate the local tangent space $T_{\mathbf{x}_i}(\mathcal{M}) \subset \mathbb{R}^d$, the linear subspace that first-order approximates $\mathcal{M}$ near $\mathbf{x}_i$ from the other representations obtained from natural inputs.

Let $\mathcal{N}_k(\mathbf{x}_i)$ be the $k$ nearest neighbors of $\mathbf{x}_i$ among the natural (unedited) representations $\mathbf{X}^{(0)}$, and let $\bar{\mathbf{x}}_{\mathcal{N}_i}$ be their mean. These neighbors provide local samples of how natural representations vary around $\mathbf{x}_i$. We form a local PCA matrix by mean-centering the neighbor positions, with one row $\mathbf{x}_j - \bar{\mathbf{x}}_{\mathcal{N}_i}$ for each neighbor $\mathbf{x}_j \in \mathcal{N}_k(\mathbf{x}_i)$. Arranging these $k$ centered local variation samples as rows gives $\mathbf{S}_i \in \mathbb{R}^{k \times d}$. We then compute SVD on $\mathbf{S}_i$ to summarize them into the dominant directions of nearby variation:
\begin{equation}
\label{eq:local_pca_svd}
\mathrm{SVD}(\mathbf{S}_i) = \mathbf{L}_i \, \mathrm{diag}(\sigma_{i,1}, \ldots, \sigma_{i,k}) \, \mathbf{V}_i^\top.
\end{equation}
The right singular vectors are the local directions in which neighbors extend, and the singular values measure how far the centered neighborhood extends along each direction. Following local tangent-space approximation methods such as LTSA \citep{zhang-zha-2004-ltsa}, we keep the top $r$ right singular vectors as a tangent-basis matrix whose column span is our estimate of $T_{\mathbf{x}_i}(\mathcal{M})$:
\begin{equation*}
\mathbf{B}_i = [\mathbf{v}_i^{(1)}, \ldots, \mathbf{v}_i^{(r)}] \in \mathbb{R}^{d \times r},
\qquad
\boldsymbol{\sigma}_i = (\sigma_{i,1}, \ldots, \sigma_{i,r}).
\end{equation*}
$\mathrm{span}(\mathbf{B}_i)$ is the estimated tangent space $T_{\mathbf{x}_i}(\mathcal{M})$. The columns of $\mathbf{B}_i$ form the basis for that subspace, so the edit is constructed by first expressing the probe's gradient in this local tangent basis and then mapping the resulting tangent-basis coordinates back to representation space. The rank $r$ is the number of tangent-basis directions retained. It should be small enough to avoid unsupported directions while large enough to capture the local geometry; when $r$ is not set directly, we estimate the intrinsic dimension of the current representations once with the TwoNN method of \citet{facco-etal-2017-two-nn} and keep that rank fixed throughout the editing loop. Although $r$ is fixed, we recompute the local tangent basis $\mathbf{B}_i$ for each sample at every erasure round. The basis is re-estimated at the sample's \emph{current} (edited) position $\mathbf{x}_i^{(t-1)}$, but the neighborhood it is estimated from is always drawn from the fixed natural representations $\mathbf{X}^{(0)}$. Thus, as a point is edited it queries a new neighborhood of the \emph{natural} representations: the tangent reflects how natural representations vary near the edited location, rather than how the already-edited points vary among themselves. This keeps the manifold estimate anchored to the manifold (the region MCH posits we should stay within) while still tracking the moving edit location.

\paragraph{Step 2: Build the tangent erasure direction.}

After estimating the local tangent basis $\mathbf{B}_i$, we turn the concept probe's gradient into a concept-erasure direction expressed in that basis. At erasure round $t$, the probe's gradient tells us how changing a representation changes the current concept probe's target prediction. For sample $i$, normalize this input gradient,
\begin{equation*}
\mathbf{u}_i
=
\frac{\nabla f_t(\mathbf{x}_i)}{\|\nabla f_t(\mathbf{x}_i)\|_2}.
\end{equation*}
Projecting onto the local tangent basis gives the vector of tangent-basis coordinates, i.e., the weights of the normalized gradient along the columns of $\mathbf{B}_i$:
\begin{equation*}
\mathbf{c}_i = \mathbf{B}_i^\top \mathbf{u}_i,
\qquad
\mathbf{B}_i\mathbf{c}_i = \mathbf{B}_i\mathbf{B}_i^\top\mathbf{u}_i.
\end{equation*}
The norm of $\mathbf{B}_i\mathbf{c}_i$ is at most one, so the update is automatically attenuated when the gradient has little support in the estimated tangent space.

We then form the \emph{spectrally-weighted tangent direction} by reweighting each tangent-basis coordinate by its singular value, giving more mass to basis directions that are well supported by nearby natural representations and less mass to thin directions with little local variation. The coordinate $c_{i,\ell}$ is the $\ell$th entry of $\mathbf{c}_i$:
\begin{equation}
\label{eq:tangent_direction}
\mathbf{d}_i
\;\triangleq\;
\mathbf{B}_i\operatorname{diag}(\boldsymbol{\sigma}_i^\alpha)\mathbf{c}_i
\;=\;
\sum_{\ell=1}^{r}\sigma_{i,\ell}^{\alpha}\,c_{i,\ell}\,\mathbf{v}_i^{(\ell)}.
\end{equation}
We use $\alpha=1$ in all the experiments; Appendix~\ref{sec:appendix_alpha_regimes} discusses the interpretation of different values of $\alpha$.

The goal of the update is to reduce the part of the current representation aligned with its local erasure direction. We therefore normalize $\mathbf{d}_i$ to $\hat{\mathbf{u}}_i = \mathbf{d}_i / \|\mathbf{d}_i\|_2$ and subtract the $\hat{\mathbf{u}}_i$-component of $\mathbf{x}_i$, scaled by the per-sample step size $\lambda_i$ chosen below:
\begin{equation}
\label{eq:tangent_erase}
\tilde{\mathbf{x}}_i
\;=\;
\mathbf{x}_i
\;-\;
\lambda_i \,\langle \mathbf{x}_i,\, \hat{\mathbf{u}}_i\rangle\, \hat{\mathbf{u}}_i .
\end{equation}
\paragraph{Step 3: Choose a per-sample local-neighborhood cap.}
\label{sec:closed_form_lambda}

A single global step size can over-edit some samples and under-edit others, so \MANCE{} determines $\lambda_i$ separately for each sample $i$. Given the tangent direction $\mathbf{d}_i$, $\lambda_i$ is chosen to make the largest bounded erasure update that stays within a local-neighborhood cap. The cap limits the displacement to an $\varepsilon$ fraction of the distance from $\mathbf{x}_i$ to its nearby representations:
\begin{equation}
\label{eq:local_radius_cap}
\|\tilde{\mathbf{x}}_i - \mathbf{x}_i\|_2 \;\le\; \varepsilon \cdot r_i .
\end{equation}
This keeps the first-order tangent approximation local: the farther the update moves from the neighborhood used to estimate $\mathbf{B}_i$, the less the local tangent basis describes the geometry around the edited point. The per-sample neighborhood radius is
\begin{equation}
\label{eq:r_i_def}
r_i \;=\; \frac{1}{k}\sum_{j\in\mathcal{N}_k(\mathbf{x}_i)} \|\mathbf{x}_j - \mathbf{x}_i\|_2 ,
\end{equation}
i.e., the mean distance from $\mathbf{x}_i$ to its $k$ nearest neighbors among the natural representations $\mathbf{X}^{(0)}$. Substituting Eq.~\ref{eq:tangent_erase} into Eq.~\ref{eq:local_radius_cap} and solving for the largest feasible $\lambda_i$ gives the closed-form per-sample step size
\begin{equation}
\label{eq:lambda_closed_form}
\lambda_i
\;=\;
\min\!\left(\;
\lambda_{\max} \;,\;
\frac{\varepsilon \cdot r_i}{\bigl|\langle \mathbf{x}_i,\,\hat{\mathbf{u}}_i\rangle\bigr|}
\;\right).
\end{equation}
Thus $\lambda_i$ is the largest step along $\hat{\mathbf{u}}_i$ that satisfies the local-neighborhood cap, subject to the hard upper bound $\lambda_{\max}$. The parameter $\varepsilon$ sets the intended displacement as a fraction of the local neighborhood radius $r_i$; if $\lambda_{\max}$ is smaller than the cap-implied step, the actual displacement is smaller. Because $r_i$ is itself a per-sample, per-panel quantity computed from the local representations, $\varepsilon$ is dimensionless in the panel's representation scale and transfers across settings without per-setting tuning (Appendix~\ref{sec:appendix_spectral_intuition}).

\begin{algorithm}[t]
\caption{\MANCE{}: Manifold aware concept erasure.}
\label{alg:mance}
\begin{algorithmic}[1]
\REQUIRE Representations $\mathbf{X}^{(0)}\in\mathbb{R}^{N\times d}$, concept labels $\mathbf{y}$, rounds $H$, neighborhood size $k$, local-neighborhood scale $\varepsilon$, probe-refit period $\tau$, cap $\lambda_{\max}$, local-basis rank $r$, spectral exponent $\alpha$.
\FOR{$t=1,\ldots,H$}
    \STATE $\displaystyle
    f_t \leftarrow
    \begin{cases}
    \text{fit MLP probe on }(\mathbf{X}^{(t-1)},\mathbf{y}),
    & t\equiv 1 \pmod \tau\\
    f_{t-1}, & \text{otherwise.}
    \end{cases}$
    \FOR{$i=1,\ldots,N$}
        \algstep{Step 1: Estimate the manifold locally.}
        \STATE \makebox[6.5cm][l]{$\displaystyle \mathcal{N}_i^{(t)} \leftarrow k\mathrm{NN}(\mathbf{x}_i^{(t-1)};\mathbf{X}^{(0)}),$}$\displaystyle \bar{\mathbf{x}}_{\mathcal{N}_i}^{(t)} \leftarrow \tfrac{1}{k}\textstyle\sum_{j\in\mathcal{N}_i^{(t)}}\mathbf{x}_j^{(0)}.$
        \STATE $\displaystyle \mathbf{S}_i^{(t)} = [\,\mathbf{x}_j^{(0)}-\bar{\mathbf{x}}_{\mathcal{N}_i}^{(t)}\,]_{j\in\mathcal{N}_i^{(t)}}.$
        \STATE $\displaystyle \mathrm{SVD}(\mathbf{S}_i^{(t)}) = \mathbf{L}_i \operatorname{diag}\bigl(\underbrace{\sigma_{i,1},\ldots,\sigma_{i,r}}_{\triangleq\ \boldsymbol{\sigma}_i},\,\ldots,\sigma_{i,k}\bigr) \mathbf{V}_i^\top.$
        \STATE $\displaystyle \mathbf{B}_i = [\mathbf{v}_i^{(1)},\ldots,\mathbf{v}_i^{(r)}].$
        \algstep{Step 2: Build the tangent erasure direction.}
        \STATE \makebox[6.5cm][l]{$\displaystyle \mathbf{u}_i = \frac{\nabla f_t(\mathbf{x}_i^{(t-1)})}{\|\nabla f_t(\mathbf{x}_i^{(t-1)})\|_2},$}$\displaystyle \mathbf{c}_i = \mathbf{B}_i^\top \mathbf{u}_i.$
        \STATE \makebox[6.5cm][l]{$\displaystyle \hat{\mathbf{u}}_i = \mathbf{d}_i \big/ \|\mathbf{d}_i\|_2,$}$\displaystyle \mathbf{d}_i = \mathbf{B}_i \operatorname{diag}(\boldsymbol{\sigma}_i^{\alpha}) \mathbf{c}_i .$
        \algstep{Step 3: Choose a per-sample local-neighborhood cap.}
        \STATE $\displaystyle r_i \;=\; \tfrac{1}{k}\textstyle\sum_{j\in\mathcal{N}_i^{(t)}} \|\mathbf{x}_j^{(0)}-\mathbf{x}_i^{(t-1)}\|_2.$
        \STATE \makebox[6.5cm][l]{$\displaystyle \mathbf{x}_i^{(t)} \leftarrow \mathbf{x}_i^{(t-1)} - \lambda_i\,\langle \mathbf{x}_i^{(t-1)},\,\hat{\mathbf{u}}_i\rangle\,\hat{\mathbf{u}}_i,$}$\displaystyle \lambda_i \;=\; \min\!\left( \lambda_{\max}, \frac{\varepsilon \cdot r_i}{\bigl|\langle \mathbf{x}_i^{(t-1)},\,\hat{\mathbf{u}}_i\rangle\bigr|} \right) .$
    \ENDFOR
\ENDFOR
\STATE \textbf{return} $\mathbf{X}^{(H)}$.
\end{algorithmic}
\end{algorithm}

\textbf{In summary,} \MANCE{} repeatedly estimates a local tangent basis around each current representation, projects the concept probe's gradient into that basis, reweights the tangent-basis coordinates by the local singular spectrum, and applies the largest per-sample update allowed by the local neighborhood radius. Algorithm~\ref{alg:mance} gives the complete editing loop. The variants below are optional preprocessing steps that run before this loop; they are not additional stages required by core \MANCE{}. 

\paragraph{\MANCEp{} \& \MANCEpp{}}

We propose two additional variants of \MANCE{} that appends another step that complements the concept removal.
Before running the \MANCE{} loop, we remove dataset-level moment structure that already has simple closed-form erasers. \MANCEp{} prepends \emph{LEACE} \citep{belrose-etal-2023-leace}, which removes first-moment linear signal associated with class means. \MANCEpp{} additionally prepends \emph{CovMatch}, a rank-2 specialisation of $k$-LEACE \citep{singh-ravfogel-2024-representation-surgery} that removes second-moment class-conditional covariance asymmetry, represented by the leading eigenvectors of $\Delta\mathbf{\Sigma} = \mathbf{\Sigma}_+ - \mathbf{\Sigma}_-$.
Both are one-shot affine projections of effective rank $\le 3$ (negligible relative to representation dimension $d \in [768, 5376]$ across our settings); implementation details for CovMatch and its preprocessing ablation are in Appendix~\ref{sec:appendix_preprocessing_derivation}.

\section{Experimental setup}
\label{sec:experimental_setup}

We use two standard evaluation metrics from the concept erasure literature, that take as input a representation, train a probe to predict some target concept, and report the performance of such probe on the test set.
\emph{Target leakage} measures the performance of the probe on target concept, and \emph{surgicality} measures the performance on a set of control concepts.

\subsection{Metrics}
\label{sec:evaluation}

\paragraph{Target leakage $S$.}
To quantify the remaining information about an erased concept, following \citet{elazar-goldberg-2018-adversarial} we train a new nonlinear probe on the representations and denote its accuracy on the test set by $S$.\footnote{Unless otherwise specified, the new probe is a 2-layer MLP with hidden size $h=128$, trained for 200 SGD steps with patience 3.} We compare $S$ with the majority-vote baseline, denoted $S_{\text{floor}}$; values near $S_{\text{floor}}$ indicate that the erased information is no longer recoverable by this probe. We denote $D_S$ as the difference between the probe accuracy and the majority-vote baseline, $D_S := S - S_{\text{floor}}$. For model selection and at-chance counts we use $|D_S|$, since erasure quality depends on distance from chance. Tables report the signed $D_S$ at the selected step only to show whether the final probe is over-erasing.

\paragraph{Surgicality $\Delta Y$.}
We measure surgicality as the damage to \emph{control concepts}. Each setting defines a control concept, and reports the test set accuracy of a trained probe on the original representations $Y_{\text{clean}}$ and on the representations $Y_{\text{edit}}$. We measure the difference between the two: $\Delta Y = Y_{\text{edit}} - Y_{\text{clean}}$, where negative values indicate control degradation; an ideal surgical edit has $\Delta Y \approx 0$.

\paragraph{Surgicality budget $D_Y$.}
To compare erasure methods fairly, we evaluate them under fixed limits on how much they may degrade control-concept accuracy. For each budget, and for each method, we choose the trajectory step that gives the strongest target erasure while still satisfying that limit.
We define the non-negative control-degradation measure
$D_Y = \max(0,\, -\Delta Y)$,
so only decreases in control accuracy count against the budget; improvements in control accuracy count as zero degradation. An edit is \emph{within budget} $b$ when $D_Y \le b$. Iterative erasers produce a sequence of edited representations, with different points along the sequence trading off target leakage against control degradation. Among the steps that are within budget, we report the step with the smallest target leakage, measured by $|D_S|$. We report budgets $D_Y \le b$ for $b \in \{1, 3, 5, 10\}$\,pp. Throughout, pp abbreviates \emph{percentage points}: all leakage ($D_S$) and surgicality ($\Delta Y$) quantities are differences between accuracies, expressed in points on the $0$--$100$ scale.

\paragraph{Coverage.}
A method need not satisfy a given budget $b$ on every setting: an eraser may exceed $b$ at all of its trajectory steps. We define \emph{coverage} as the number of settings, out of the total $N$, on which a method has at least one within-budget edit, reported as $n/N$. Because $|D_S|$ is averaged only over covered settings, coverage and leakage must be read together: a small mean $|D_S|$ over few covered settings is weaker evidence of erasure than the same value over all of them.

\subsection{Settings, methods, and hyperparameters}
\label{sec:hp_selection}

\paragraph{Settings and probe.} We perform extensive evaluations of our methods and previous baselines on 119 settings: 39 NLP settings (13 LLM families $\times$ 3 concepts at the 50\%-depth layer of each model) and 80 CelebA-CLIP settings (CLIP ViT-B/32 \citep{radford-etal-2021-clip} final-layer pooled image embeddings of CelebA \citep{liu-etal-2015-celeba}; 40 binary facial attributes $\times$ 2 surgicality regimes). The 13 LLMs span 0.5B–27B parameters: Qwen2.5 (0.5B, 1.5B, 3B), Gemma-2 (2B, 9B, 27B), Gemma-3 (1B, 4B, 12B, 27B), Llama-3.2 (1B, 3B), Mistral-7B-v0.1. 
The three NLP concepts are \textit{sycophancy} \citep{perez-etal-2022-model-written}, \textit{gender} (Bias in Bios \citep{dearteaga-etal-2019-bias}), and \textit{safety} (PKU-SafeRLHF \citep{ji-etal-2025-pku-saferlhf}). 
The control concepts for these three datasets are answer preference for sycophancy, profession for gender, and helpfulness for safety, respectively. In PKU-SafeRLHF, examples are response pairs; we use the dataset's helpfulness preference within each pair to label the preserved control concept, while the safety annotation defines the erasure target.

For CelebA, we use each of the 40 attributes as the erasure target under two surgicality regimes: \textit{highly entangled} and \textit{highly disentangled} control concepts. These two regimes are formed from the five attributes
  most and least correlated with the target concept on the training split, respectively. 
For example, when erasing \emph{Male}, the most-correlated controls are \emph{Wearing\_Lipstick} ($|r| \approx 0.80$), \emph{Heavy\_Makeup}, \emph{No\_Beard}, \emph{Attractive}, and \emph{5\_o\_Clock\_Shadow}. The resulting control sets, with their $|r|$ values, are listed in App.~Tabs.~\ref{tab:celeba_controls_least} and~\ref{tab:celeba_controls_most}; additional construction details are in App.~\ref{sec:appendix_celeba_controls}.

\paragraph{Baselines.} In addition to the \MANCE{} family, we compare against several baselines. LEACE \citep{belrose-etal-2023-leace} and LEACE\,+\,CovMatch are closed-form one-shot linear and rank-2 covariance erasers. INLP \citep{ravfogel-etal-2020-null} is an iterative linear nullspace projection method. IGBP \citep{iskander-etal-2023-shielded} is an iterative 2-class probe with Newton-step boundary projection. Obliviator \citep{akbari-etal-2025-obliviator} is the current state-of-the-art nonlinear method, which iteratively trains an encoder to minimize statistical dependence between the representation and the erased attribute while preserving task-relevant information, and then applies an RKHS-based disentanglement step to refine the representation for the next round.

\paragraph{Ablation.} To isolate the value of the tangent constraint, we introduce \AmbCEpp{}, a controlled variant of \MANCEpp{} that uses the same preprocessing and nonlinear probe loop but removes the tangent projection. Instead of projecting the concept-probe gradient into the local tangent basis and scaling the step by the per-sample local-radius cap, \AmbCEpp{} takes a full-gradient step in the full representation space with a single global step size, $\lambda=29.31$. This value is the mean per-sample $\lambda_i$ produced by \MANCEpp{} on the NLP settings (App.~Tab.~\ref{tab:lambda_stats}), making the ablation comparable in effective step scale. Thus, \AmbCEpp{} tests whether the gains come from the tangent constraint rather than from the nonlinear probe loop alone, directly testing the MCH prediction (\S\ref{sec:mch}).

\paragraph{Evaluation protocol.} 
For every setting we fit the concept eraser on the training representations and evaluate each trajectory step on the held-out test split, reporting both target leakage $S$ and surgicality $\Delta Y$.
The NLP datasets do not share a standard split, we therefore impose one uniform protocol: we pool the examples, take a stratified subsample of $12{,}000$, and draw a stratified $60/20/20$ split, giving $7{,}200$ train / $2{,}400$ val / $2{,}400$ test per setting.
CelebA already provides an official train/val/test partition, which we keep, subsampling within each to $1{,}200$ / $300$ / $300$ images per setting. 
In all cases the eraser is fit on train, the val split only early-stops the probes, and both $S$ and $Y$ are reported on the held-out test split.
The new nonlinear MLP defined in \Cref{sec:evaluation} measures target leakage $S$, and an independent control evaluator measures $Y$ on the corresponding control task or control attributes. For the NLP settings, each dataset comes with the target and control concepts. 
For CelebA, we compute $S$ on each of the 40 attributes, and $Y$ as the mean accuracy on the regime's five control attributes. 
Across both modalities, we report the four \textit{surgicality budgets} $D_Y \in \{1, 3, 5, 10\}$pp, meaning average control doesn't drop below the budget relative to the clean (un-erased) representation.

\paragraph{Hyperparameters.} All step-controlling hyperparameters are fixed across the 119 settings with no per-setting tuning; values are reported in App.~\ref{sec:appendix_hyperparameters}.

\section{Results}
\label{sec:results}

We evaluate the \MANCE{} algorithms family (\S\ref{sec:method}), and previous erasure methods as baselines.
We test all methods on 119 settings: 39 from NLP, spanning 13 language models and three concepts, plus 80 from vision, using CelebA with 40 image attributes (the concepts we erase in the vision setting) under two surgicality regimes defined by target--control label correlation: \textit{high-correlation} (5 most-correlated controls) and \textit{low-correlation} (5 least-correlated controls). We use this correlation as a measurable proxy for representational entanglement, a relationship correlation is expected to induce \citep{elhage2022toy}.
For iterative methods (e.g., INLP), we run the evaluation metrics for each iteration.
Then, per surgicality budget, we select the iteration that erases the most of the target concept.
In the non-iterative methods, we simply use the obtained, single result.

\paragraph{Adding \MANCE{} improves prior erasers.}
Applying \MANCE{} on intervened representations from other erasers leads to improved results across most settings.
\Cref{tab:mance_on_baseline} reports leakage gains obtained by applying \MANCE{} after each erasure method, aggregated over the $39$ NLP settings ($13$ LLMs $\times$ three concepts: sycophancy, gender, and safety; see \S\ref{sec:hp_selection}).
Under the same surgicality budgets, adding \MANCE{} drives each baseline's target leakage close to chance: at $D_Y \le 1$pp it falls from $19.1$ to $1.5$pp for LEACE, $15.2$ to $1.8$pp for INLP, and $11.5$ to $1.6$pp for IGBP, with similar drops at the looser budgets, and it keeps as many or more settings within budget (\Cref{tab:mance_on_baseline}).
The only exception is Obliviator, where we see no improvement in leakage. However, Obliviator's coverage (\S\ref{sec:evaluation}) is only $13$--$19/39$; on those few covered settings its leakage is already down to random, so \MANCE{} has nothing left to remove.
This apparent strength is an artifact of \emph{which} settings it reaches. 
Next, we will show how for harder, highly correlated settings, \MANCEpp{} outperforms Obliviator (and other baselines).

\begin{table}[t]
\centering
\caption{\textbf{\MANCE{} complements prior erasers.} Mean residual leakage $|D_S|$ (pp); \emph{lower is better in both columns}. \emph{Alone} ($\downarrow$): the baseline's leakage. \emph{$+$\MANCE{}} ($\downarrow$): leakage after additionally applying \MANCE{} to the baseline's representations. Parentheses give each method's \emph{coverage} in that column (\S\ref{sec:evaluation}), the number of the $39$ settings it keeps within budget; every mean is taken only over those covered settings, so the two columns also show that \MANCE{} keeps at least as many settings within budget. \MANCE{} on top of LEACE and of LEACE\,+\,CovMatch recovers our \MANCEp{} and \MANCEpp{} variants (noted in the row labels). $^{\dagger}$Obliviator's $0.0$ entries cover only the $13$--$19/39$ settings it keeps within budget, on which it is already at chance, so \MANCE{} has nothing to remove; on the entangled settings it violates the surgicality budget and reports no result (\Cref{tab:main_results_unified}).}
\label{tab:mance_on_baseline}
\footnotesize
\setlength{\tabcolsep}{2pt}
\resizebox{\textwidth}{!}{%
\begin{tabular}{l|cc|cc|cc|cc}
\toprule
& \multicolumn{2}{c|}{$D_Y \le 1$pp} & \multicolumn{2}{c|}{$D_Y \le 3$pp} & \multicolumn{2}{c|}{$D_Y \le 5$pp} & \multicolumn{2}{c}{$D_Y \le 10$pp} \\
Method & Alone$\downarrow$ & $+$\MANCE{}$\downarrow$ & Alone$\downarrow$ & $+$\MANCE{}$\downarrow$ & Alone$\downarrow$ & $+$\MANCE{}$\downarrow$ & Alone$\downarrow$ & $+$\MANCE{}$\downarrow$ \\
\midrule
\MANCE{} (no preprocessing) & -- & $4.5$~(37/39) & -- & $2.4$~(38/39) & -- & $2.2$~(39/39) & -- & $1.7$~(39/39) \\
INLP & $15.2$~(36/39) & $1.8$~(38/39) & $15.6$~(37/39) & $1.5$~(39/39) & $16.0$~(39/39) & $1.1$~(39/39) & $16.0$~(39/39) & $1.0$~(39/39) \\
IGBP & $11.5$~(38/39) & $1.6$~(38/39) & $11.5$~(39/39) & $0.9$~(39/39) & $11.5$~(39/39) & $0.7$~(39/39) & $11.5$~(39/39) & $0.6$~(39/39) \\
Obliviator & $0.0^{\dagger}$~(13/39) & $0.0^{\dagger}$~(13/39) & $0.0^{\dagger}$~(13/39) & $0.0^{\dagger}$~(13/39) & $0.0^{\dagger}$~(13/39) & $0.0^{\dagger}$~(13/39) & $0.0^{\dagger}$~(19/39) & $0.0^{\dagger}$~(19/39) \\
LEACE {\footnotesize($+$\MANCE{}${=}$\MANCEp{})} & $19.1$~(38/39) & $1.5$~(38/39) & $19.0$~(39/39) & $1.0$~(39/39) & $19.0$~(39/39) & $0.7$~(39/39) & $19.0$~(39/39) & $0.6$~(39/39) \\
LEACE\,+\,CovMatch {\footnotesize($+$\MANCE{}${=}$\MANCEpp{})} & -- & $1.7$~(33/39) & -- & $0.1$~(37/39) & -- & $0.1$~(38/39) & -- & $0.0$~(39/39) \\
\bottomrule
\end{tabular}%
}
\end{table}

\paragraph{\MANCEpp{} gives the strongest leakage--surgicality tradeoff.}
\label{sec:multi_model}

\begin{figure}[t]
\centering
\begin{subfigure}[t]{0.46\linewidth}
\centering
\includegraphics[width=\linewidth]{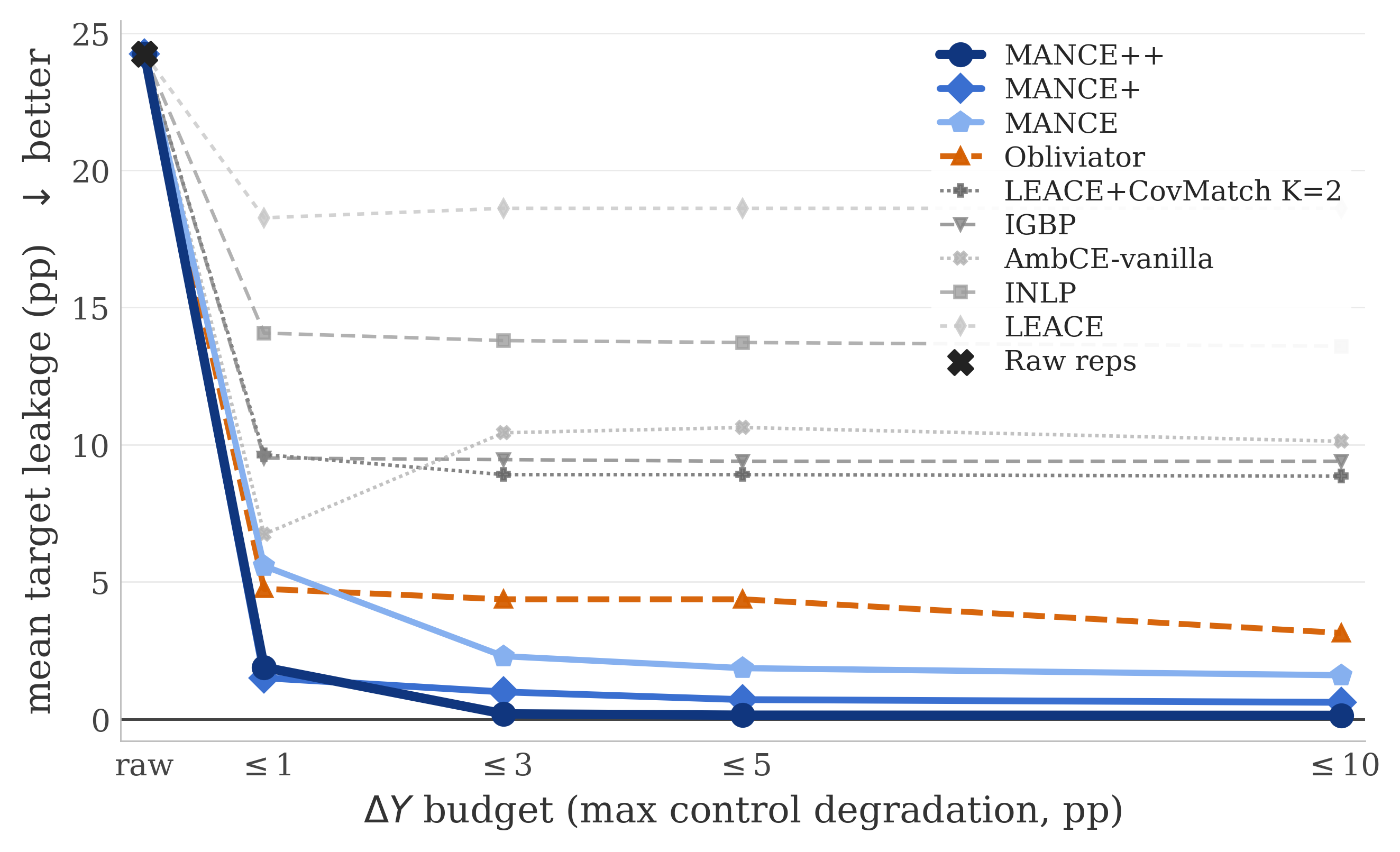}
\caption{Mean NLP leakage envelope across 39 settings; \MANCE{} variants form the lowest curves and \MANCEpp{} is below 1pp from $D_Y \le 3$pp onward.}
\label{fig:aggregated_nlp_results}
\end{subfigure}\hfill
\begin{subfigure}[t]{0.52\linewidth}
\centering
\includegraphics[width=\linewidth]{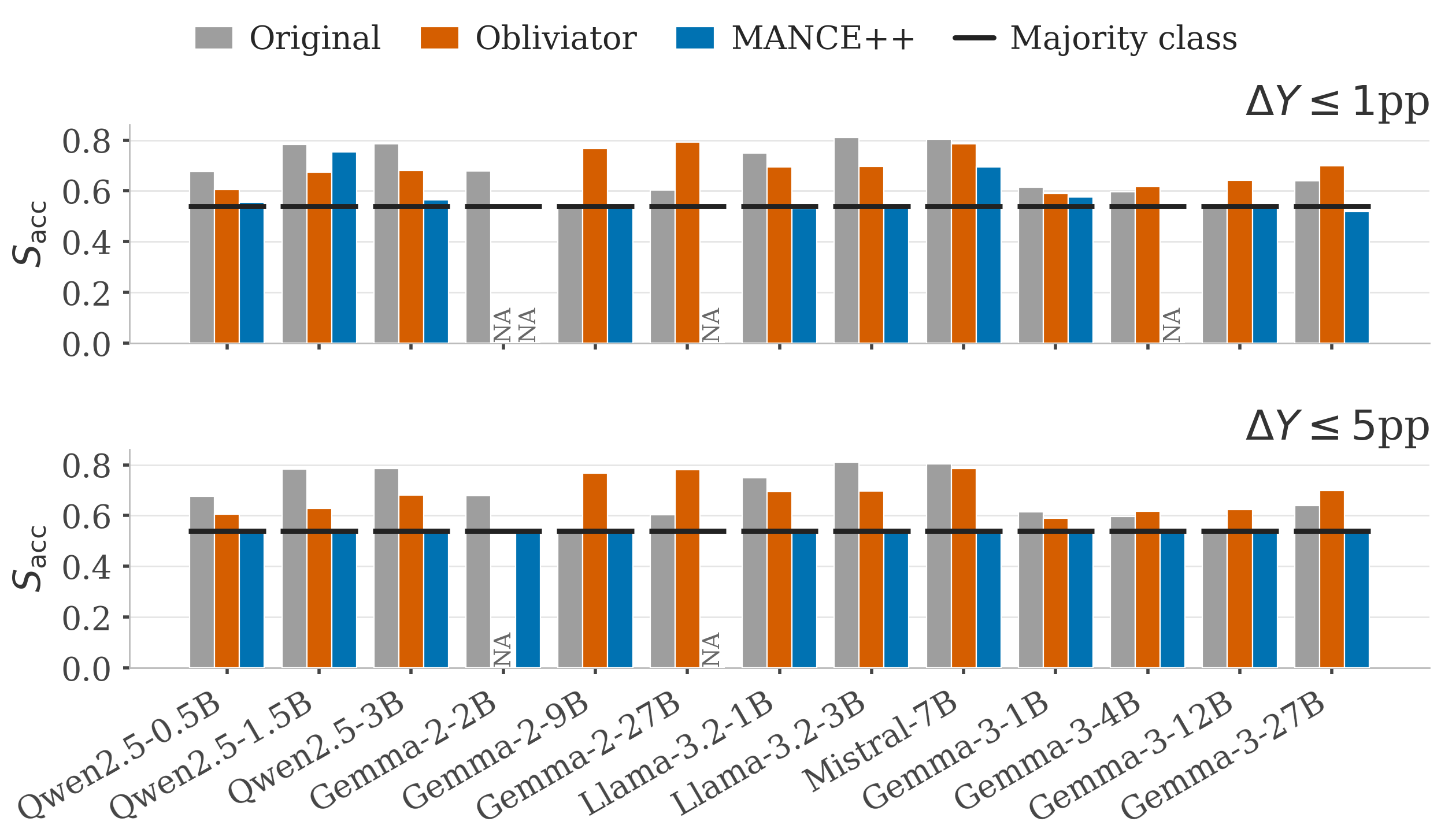}
\caption{Per-model gender nonlinear-probe accuracy at $D_Y \le 1$pp (top) and $D_Y \le 5$pp (bottom); black ticks mark majority class. \MANCEpp{} reaches chance in more settings than Obliviator.}
\label{fig:per_model_bars_gender_compact}
\end{subfigure}
\caption{\textbf{NLP leakage summaries.} (a) Mean leakage across 39 NLP settings. (b) Per-model gender accuracy at two surgicality budgets.}
\label{fig:nlp_results_summary}
\end{figure}

\begin{figure}[t]
\centering
\includegraphics[width=\linewidth]{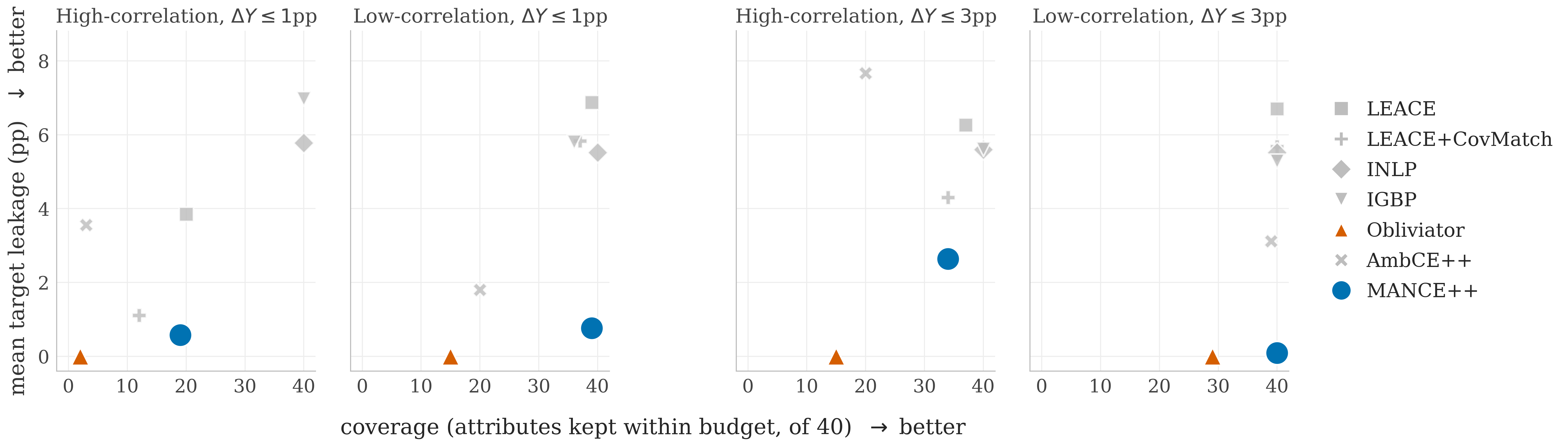}
\caption{\textbf{Vision (CelebA): coverage vs.\ mean target leakage.} Each point is one erasure method: $x$-axis is coverage (how many of the $40$ CelebA attributes it erases within the surgicality budget), $y$-axis is its mean target leakage over those covered attributes; lower-right is better. Panels pair the two budgets ($D_Y\le 1$pp, $D_Y\le 3$pp) with the high- and low-correlation control regimes. 
\MANCEpp{} shows consistent strong tradeoffs between target leakage and coverage, and marks a new pareto curve.
Obliviator (orange) matches its low leakage only at far lower coverage, while LEACE, INLP, and IGBP cover many attributes but leave $4$--$7$pp leakage.
}
\label{fig:celeba_dual_regime}
\end{figure}

\Cref{tab:main_results_unified,fig:nlp_results_summary} present the NLP comparison and \Cref{fig:celeba_dual_regime} the vision comparison; complete per-setting results appear in the appendix (NLP in \Cref{tab:per_panel_breakdown,fig:per_model_bars_gender_full,fig:per_model_bars_sycophancy_app,fig:per_model_bars_safety_app}, vision in \Cref{tab:celeba_unified_summary,fig:per_attr_bars_celeba}). Both modalities show the same trend: \MANCEpp{} outperforms all baselines, setting a new state of the art in nonlinear concept erasure.
On the NLP settings, \MANCEpp{} is the only method that stays near chance at every budget: $+1.6 \to 0.0$pp from $\le 1$pp to $\le 10$pp, reaching chance on $19$--$35/39$ settings. On the other hand, the strongest baseline, Obliviator, retains $+4.3 \to +2.7$pp above the majority-vote and reaches chance on $13$--$17/39$.
In the vision setting (\Cref{fig:celeba_dual_regime}), under the same configuration and preprocessing as on NLP, \MANCEpp{} is the only method that combines high coverage with near-floor leakage. The high-coverage baselines (LEACE, INLP, and IGBP) reach most or all attributes but leave $4$--$7$pp of leakage; the only other near-floor method, Obliviator, instead attains far lower coverage. 

In the least-correlated regime, \MANCEpp{} covers $39/40$ and $40/40$ attributes at $D_Y \le 1 pp$ and $D_Y \le 3 pp$, respectively, compared with $15/40$ and $29/40$ for Obliviator. In the most-correlated regime, \MANCEpp{} covers $19/40$ and $34/40$ attributes, respectively, compared with $2/40$ and $15/40$ for Obliviator.
Obliviator's near-floor leakage in \Cref{fig:celeba_dual_regime} is biased by coverage: it is averaged only over the easy attributes Obliviator stays in budget, where \MANCEpp{} also reaches the floor. Because \MANCEpp{} stays within the surgicality budget on the harder attributes too, it additionally reaches near-chance erasure there, whereas Obliviator has no in-budget result to report (complete per-attribute values in \Cref{tab:celeba_unified_summary,fig:per_attr_bars_celeba}). Thus, across text and vision, tangent-based erasure is more effective under the same surgicality budgets, providing empirical support for MCH (Sec.~\ref{sec:mch}).

\begin{table}[t]
\centering
\caption[Main NLP erasure result]{\textbf{Main NLP erasure result.} We report two numbers per surgicality budget. \emph{Leakage} ($\downarrow$): how much target information survives, measured as $D_S$ in pp at the within-budget step closest to chance, so $0$ means the probe is at majority-class chance and larger values mean more leakage. \emph{At Chance} ($\uparrow$), written $n/d$: of the $d$ settings a method keeps within budget (its \emph{coverage}; \S\ref{sec:evaluation}), the number $n$ it drives to chance ($|D_S| \le 0.5$pp). Per-panel detail is in App.~Tab.~\ref{tab:per_panel_breakdown}.}
\label{tab:main_results_unified}
\footnotesize
\setlength{\tabcolsep}{4pt}
\begin{tabular}{l|cc|cc|cc|cc}
\toprule
& \multicolumn{2}{c|}{$D_Y \le 1$pp} & \multicolumn{2}{c|}{$D_Y \le 3$pp} & \multicolumn{2}{c|}{$D_Y \le 5$pp} & \multicolumn{2}{c}{$D_Y \le 10$pp} \\
Method & Leakage$\downarrow$ & Chance$\uparrow$ & Leakage$\downarrow$ & Chance$\uparrow$ & Leakage$\downarrow$ & Chance$\uparrow$ & Leakage$\downarrow$ & Chance$\uparrow$ \\
\midrule
LEACE & $+18.3$ & $5/38$ & $+18.6$ & $5/39$ & $+18.6$ & $5/39$ & $+18.6$ & $5/39$ \\
LEACE$+$CovMatch & $+9.6$ & $12/30$ & $+8.9$ & $16/38$ & $+8.9$ & $16/38$ & $+8.9$ & $16/39$ \\
INLP & $+14.1$ & $7/38$ & $+13.8$ & $7/39$ & $+13.7$ & $8/39$ & $+13.6$ & $8/39$ \\
IGBP & $+9.5$ & $4/37$ & $+9.4$ & $4/38$ & $+9.4$ & $4/38$ & $+9.4$ & $4/38$ \\
Obliviator & $+4.3$ & $13/39$ & $+4.0$ & $13/39$ & $+4.0$ & $13/39$ & $+2.7$ & $17/39$ \\
\AmbCEpp{} & $+5.7$ & $7/20$ & $+9.8$ & $11/30$ & $+10.0$ & $11/32$ & $+9.8$ & $12/33$ \\
\MANCE{} & $+5.3$ & $13/35$ & $+2.0$ & $20/39$ & $+1.8$ & $24/39$ & $+1.6$ & $28/39$ \\
\MANCEp{} & $\mathbf{+1.4}$ & $23/38$ & $+1.0$ & $25/39$ & $+0.7$ & $29/39$ & $+0.6$ & $30/39$ \\
\MANCEpp{} & $+1.6$ & $19/33$ & $\mathbf{0.0}$ & $32/37$ & $\mathbf{0.0}$ & $34/38$ & $\mathbf{0.0}$ & $35/39$ \\
\bottomrule
\end{tabular}
\end{table}

\paragraph{The gain is largest where the target is hardest to remove surgically.}
\label{sec:celeba}
\MANCEpp{}'s advantage over the baselines is largest where the target is hardest to remove without damaging the control concepts. In NLP this is clearest on gender, where profession strongly correlates with gender: \MANCEpp{} reaches chance on $3/13$ models versus $0/13$ for Obliviator at $D_Y \le 1$pp, widening sharply to $12/13$ versus $0/13$ at $D_Y \le 5$pp (\cref{fig:per_model_bars_gender_compact}; App.~\cref{fig:per_model_bars_gender_full,tab:per_panel_breakdown}). Safety is intermediate: \MANCEpp{} reaches chance on $11/13$ models, while Obliviator overshoots the budget on $2/13$ settings at $D_Y \le 1$pp (App.~\cref{fig:per_model_bars_safety_app}). On the other hand, Sycophancy is already solved by linear erasure methods, leaving nothing for \MANCE{} to add (App.~\cref{fig:per_model_bars_sycophancy_app,tab:leace_vs_mance_nlp}). Finally, the vision setting shows the same pattern, where the difficulty is measured directly as target--control correlation: under the most-correlated controls \MANCEpp{} keeps far more attributes within budget than Obliviator while staying near the target leakage floor (\Cref{fig:celeba_dual_regime}).

\paragraph{The manifold constraint makes erasure precise and surgical.}

The unconstrained ablation \AmbCEpp{} matches \MANCEpp{} but takes full-space gradient steps of the same effective magnitude (the average $\lambda$ over \MANCEpp{}'s iterations, $\lambda{=}29.31$; App.~Tab.~\ref{tab:lambda_stats}). Without the manifold constraint it leaves $6$--$10$pp leakage, and on many settings no step of its trajectory stays within the surgicality budget; by the coverage protocol (\S\ref{sec:evaluation}) those out-of-budget settings are excluded from its leakage mean, so even that $6$--$10$pp is measured only over the easier settings it keeps in budget. Compared with \MANCEpp{}'s $+1.6 \to 0.0$pp, the gains come from the manifold constraint, not the closed-form preprocessing or the nonlinear probe loop alone.

\paragraph{The latency tradeoff.}
We benchmark the latency of the different methods on the Gender concept.
\MANCE{} takes about eight minutes ($458.8$--$474.9$\,s across the three variants on one NVIDIA~B200), with the closed-form preprocessing in \MANCEp{}/\MANCEpp{} adding only a few seconds on top; the baselines range from a few seconds (LEACE, CovMatch) to a few minutes (INLP) (\Cref{tab:method_latency}, \S\ref{sec:appendix_latency}).
Profiling attributes roughly $50\%$ of the runtime to the per-round local SVDs and $40\%$
to CPU--GPU transfers and CPU-side projection arithmetic. Both are properties of the
current implementation rather than of the algorithm, and a fully GPU-resident
implementation would reduce the runtime severalfold.

\paragraph{Summary.} Across the 119 settings, we observe consistent results: 
Applied to the representations produced by previous erasure methods, \MANCE{} further improves erasure at equal or better coverage under the same surgicality budget.
In addition, our \MANCEpp{} variant achieves state-of-the-art concept-erasure results across various models, datasets, and modalities.
Our results provide strong empirical support for the Manifold Constraint Hypothesis (\S\ref{sec:mch}): constraining edits to locally supported directions improves surgical concept erasure.

\section{Discussion}

We believe MCH's effectiveness go beyond concept erasure: interventions on representations should respect the geometry that the model and the data induce jointly, rather than unconstrained modifications. In this work, we study erasure, but the same argument should apply to other representation editing such as activation steering \citep{subramani-etal-2022-extracting, turner2023steering}.

Our method is limited by how well we can estimate $\mathcal{M}$. \MANCE{} uses a local, first-order estimate, a per-representation tangent space that is cheap and transfers across models and modalities without per-setting tuning. We believe that better manifold estimators, including non-local and more global models of $\mathcal{M}$, would enable more surgical interventions and let one estimate serve different edits instead of being recomputed for each representation. Improving manifold estimation, and relating its quality to guarantees on interventions, is in our view a promising path toward geometry-aware interventions.

\section{Conclusion}

We proposed the \emph{Manifold Constraint Hypothesis}: if natural representations concentrate on a manifold, then interventions constrained to the manifold should preserve other information encoded in the representation. 
\MANCE{} instantiates this hypothesis for concept erasure: it estimates the manifold from neighboring representations, follows the nonlinear concept probe toward reduced leakage, and confines the update to the manifold.

We perform extensive experiments using 119 settings spanning text and vision models and datasets: three NLP concepts (sycophancy, gender, safety) tested on 13 language models, and 40 CelebA-CLIP attributes.
\MANCE{} improves leakage--surgicality tradeoffs relative to unconstrained updates and improves prior erasure methods under the same surgicality budget. These results support the Manifold Constraint Hypothesis which is useful for erasure, while leaving better manifold estimators and formal links between tangent-estimation quality and erasure guarantees for future work.

\section*{Limitations}

\paragraph{Measurement scope.} We quantify erasure by retraining a nonlinear MLP probe on the edited representations, following the recoverability protocol of \citet{elazar-goldberg-2018-adversarial}, and we quantify preservation by accuracy on the labeled control concepts. Both target leakage $S$ and surgicality $\Delta Y$ are thus empirical measurements under a fixed probe protocol rather than guarantees that the target concept is fully unrecoverable. Surgicality in particular speaks only to the control concepts we enumerated (Sec.~\ref{sec:hp_selection}), not to the full information content of the representation.

\paragraph{Local manifold estimation.} \MANCE{} constrains each edit to a local tangent space $T_{\mathbf{x}_i}(\mathcal{M})$ estimated by local PCA over the $k$ nearest natural representations $\mathbf{X}^{(0)}$, with the tangent rank $r$ taken from a TwoNN intrinsic-dimension estimate. This estimate degrades where the natural representations $\mathbf{X}^{(0)}$ are sparse or the manifold is strongly curved. The tangent constraint also helps only to the extent that MCH holds, that is, that representations concentrate on a structured, lower-dimensional manifold, so its advantage over full-space editing is expected to shrink as the intrinsic dimension approaches the representation dimension $d$.

\paragraph{Computational cost.} Unlike closed-form linear erasers such as LEACE, which apply a single affine map, \MANCE{} recomputes a $k$-NN query and a local SVD for every representation at every round and periodically refits the nonlinear probe. Applying the fitted eraser to new inputs likewise requires querying the natural representations $\mathbf{X}^{(0)}$ for their nearest neighbors, so both fitting and deployment are heavier than one-shot projection methods.

\bibliographystyle{plainnat}
\bibliography{references}

\appendix

\section{Closed-form preprocessing details}
\label{sec:appendix_preprocessing_derivation}

The two \MANCE{} preprocessing variants (\MANCEp{} and \MANCEpp{}) exhaust 1st- and 2nd-moment concept signal in closed form so that the subsequent manifold-constrained loop only has to attack the residual nonlinear pocket.

\paragraph{LEACE (1st moment).}
Fit a LEACE eraser \citep{belrose-etal-2023-leace} on $(\mathbf{X}_{\text{train}}, \mathbf{y})$ and apply to all splits. LEACE removes the rank-1 cross-covariance $\mathbf{\Sigma}_{xz}$, leaving no \emph{linear} classifier above majority-class accuracy with provably minimum representation damage among affine projections.

\paragraph{Covariance-asymmetry projection (2nd moment).}
LEACE matches class means but leaves the within-class asymmetry
\begin{equation}
\label{eq:delta_sigma}
\Delta\mathbf{\Sigma} = \mathbf{\Sigma}_{+} - \mathbf{\Sigma}_{-}.
\end{equation}
A nonlinear probe still recovers concept information from variance asymmetry along axes where $\mathbf{\Sigma}_{+}$ and $\mathbf{\Sigma}_{-}$ differ. We append a rank-$2$ projection: the top-$2$ eigenvectors of $\Delta\mathbf{\Sigma}$ ranked by $|\lambda|$, orthonormalized with a residual mean direction via QR,
\begin{equation}
\label{eq:covmatch_basis}
\mathbf{D} = \mathrm{QR}\!\left([\,\hat{\mathbf{d}}_{\text{mean}} \;\big|\; \mathbf{e}_1, \mathbf{e}_2\,]\right) \in \mathbb{R}^{d \times 3},
\end{equation}
\begin{equation}
\label{eq:covmatch_apply}
\tilde{\mathbf{x}} = \mathbf{x} - \mathbf{D}\mathbf{D}^\top \mathbf{x}.
\end{equation}
Because LEACE has aligned class means, $\hat{\mathbf{d}}_{\text{mean}} \approx \mathbf{0}$ and the effective rank removed is two. This is a special case of the $k$-LEACE framework of \citet{singh-ravfogel-2024-representation-surgery} restricted to second-order eigenvectors, which are stable to estimate from $n \approx 7000$ examples while higher-order tensors are not. We keep the top two eigenvectors across all 119 settings: for a binary concept, these directions capture the two axes of class-conditional variance asymmetry; subsequent eigenvectors describe noise rather than class-discriminative signal. Empirically, keeping three or more eigenvectors adds no leakage reduction and begins to erode control accuracy (see Tab.~\ref{tab:per_panel_breakdown}).

The two stages erase structurally distinct signal (mean shift vs.\ variance asymmetry) and together remove an effective rank of at most $3$, negligible relative to the representation dimension $d$ across our settings ($d = 768$ for the CLIP ViT-B/32 image embedding, and $d \in [896, 5376]$ across the 13 LLM hidden sizes), exposing the residual nonlinear structure that the manifold-constrained loop attacks.

\section{Closed-form preprocessing ablation}
\label{sec:appendix_preprocessing_ablation}

Tab.~\ref{tab:leace_vs_mance_nlp} reports a new nonlinear probe ($h{=}128$, 200 SGD steps, patience 3) trained on three representation states per setting: clean, post-LEACE, post-(LEACE\,+\,CovMatch). Sycophancy is fully exhausted by closed-form preprocessing (mean NL $0.563$ indistinguishable from floor $0.560$ and from \MANCEpp{}'s $0.560$). Gender is not: LEACE\,+\,CovMatch leaves $14.8$pp residual above floor that the manifold-constrained loop closes to $1.0$pp. Safety follows the gender pattern with $11.5 \to 0.2$pp.

\begin{table}[t]
\centering
\small
\caption[Closed-form preprocessing ablations]{\textbf{Closed-form preprocessing ablations on the 39-panel NLP grid (50\%-depth layer).} A new nonlinear probe ($h=128$, 200 SGD steps, patience 3) is trained on each post-erasure representation. \emph{LEACE} is the affine LEACE eraser. \emph{LEACE\,+\,CovMatch} additionally projects out the top-2 eigenvectors of $\Delta\Sigma = \Sigma_{+} - \Sigma_{-}$ (rank-3 closed-form erasure total). \emph{\MANCEpp{}} is the proposed full pipeline (LEACE + CovMatch preprocessing + iterative manifold-constrained loop, $\varepsilon{=}0.1$); we report the best $|S{-}\text{floor}|$ trajectory step subject to $D_Y \le 5$pp. NL = nonlinear probe accuracy on test (lower is better; floor is the per-panel majority class). $\Delta Y$ is the signed control-task change in pp. Bold marks the row's NL minimum.}
\label{tab:leace_vs_mance_nlp}
\begin{tabular}{l l c | c c | c c | c c}
\toprule
 &  & Floor & \multicolumn{2}{c|}{LEACE} & \multicolumn{2}{c|}{LEACE\,+\,CovMatch} & \multicolumn{2}{c}{\MANCEpp{} $\varepsilon{=}0.1$} \\
Model & Concept & (NL$^*$) & NL & $\Delta Y$pp & NL & $\Delta Y$pp & NL & $\Delta Y$pp \\
\midrule
Qwen2.5-0.5B & Sycophancy & 0.550 & \textbf{0.550} & 0.0 & \textbf{0.550} & -2.1 & \textbf{0.550} & -0.2 \\
 & Gender & 0.539 & 0.709 & -0.1 & 0.684 & -0.4 & \textbf{0.538} & -2.8 \\
 & Safety & 0.510 & 0.752 & 0.0 & 0.529 & 0.0 & \textbf{0.510} & 0.0 \\
\addlinespace[2pt]
Qwen2.5-1.5B & Sycophancy & 0.550 & \textbf{0.550} & -0.1 & \textbf{0.550} & -1.5 & \textbf{0.550} & -0.5 \\
 & Gender & 0.539 & 0.822 & -0.3 & 0.788 & -1.7 & \textbf{0.539} & -3.0 \\
 & Safety & 0.510 & 0.759 & +2.9 & \textbf{0.510} & +2.9 & \textbf{0.510} & +1.1 \\
\addlinespace[2pt]
Qwen2.5-3B & Sycophancy & 0.675 & \textbf{0.675} & +0.1 & \textbf{0.675} & -2.7 & \textbf{0.675} & -0.9 \\
 & Gender & 0.539 & 0.857 & -1.3 & 0.783 & -0.1 & \textbf{0.540} & -2.4 \\
 & Safety & 0.510 & 0.718 & +0.5 & 0.520 & +0.5 & \textbf{0.510} & 0.0 \\
\addlinespace[2pt]
Gemma-2-2B & Sycophancy & 0.543 & 0.643 & 0.0 & \textbf{0.543} & -0.4 & \textbf{0.543} & -0.1 \\
 & Gender & 0.539 & 0.722 & +0.3 & 0.681 & -0.7 & \textbf{0.539} & -1.7 \\
 & Safety & 0.510 & 0.715 & +0.7 & 0.654 & +0.7 & \textbf{0.510} & -0.1 \\
\addlinespace[2pt]
Gemma-2-9B & Sycophancy & 0.569 & 0.751 & -0.1 & \textbf{0.569} & -0.7 & 0.569 & +0.5 \\
 & Gender & 0.539 & 0.638 & +0.2 & 0.562 & -0.4 & \textbf{0.539} & +0.3 \\
 & Safety & 0.510 & 0.623 & 0.0 & \textbf{0.510} & 0.0 & \textbf{0.510} & 0.0 \\
\addlinespace[2pt]
Gemma-2-27B & Sycophancy & 0.553 & 0.695 & -0.2 & \textbf{0.553} & -0.9 & \textbf{0.553} & -3.3 \\
 & Gender & 0.539 & 0.663 & 0.0 & 0.605 & -2.5 & \textbf{0.539} & -3.1 \\
 & Safety & 0.510 & 0.736 & +0.7 & 0.643 & +1.4 & \textbf{0.510} & -0.9 \\
\addlinespace[2pt]
Llama-3.2-1B & Sycophancy & 0.551 & 0.849 & -0.1 & \textbf{0.551} & -0.2 & \textbf{0.551} & -0.1 \\
 & Gender & 0.539 & 0.855 & -0.4 & 0.738 & +0.8 & \textbf{0.540} & -3.4 \\
 & Safety & 0.510 & 0.817 & +0.3 & 0.778 & +0.3 & \textbf{0.511} & 0.0 \\
\addlinespace[2pt]
Llama-3.2-3B & Sycophancy & 0.551 & 0.896 & 0.0 & \textbf{0.551} & 0.0 & \textbf{0.551} & 0.0 \\
 & Gender & 0.539 & 0.889 & +0.2 & 0.794 & -0.4 & \textbf{0.539} & -0.6 \\
 & Safety & 0.510 & 0.837 & +0.1 & 0.821 & +0.2 & \textbf{0.527} & 0.0 \\
\addlinespace[2pt]
Mistral-7B & Sycophancy & 0.551 & 0.920 & 0.0 & \textbf{0.551} & -0.1 & \textbf{0.551} & 0.0 \\
 & Gender & 0.539 & 0.851 & -0.4 & 0.815 & -0.3 & \textbf{0.539} & -2.8 \\
 & Safety & 0.510 & 0.825 & 0.0 & 0.811 & 0.0 & \textbf{0.509} & 0.0 \\
\addlinespace[2pt]
Gemma-3-1B & Sycophancy & 0.551 & 0.727 & -0.1 & \textbf{0.551} & -2.2 & \textbf{0.551} & +0.5 \\
 & Gender & 0.539 & 0.708 & 0.0 & 0.613 & -0.8 & \textbf{0.539} & +1.3 \\
 & Safety & 0.510 & 0.726 & 0.0 & 0.723 & -1.0 & \textbf{0.510} & 0.0 \\
\addlinespace[2pt]
Gemma-3-4B & Sycophancy & 0.551 & \textbf{0.551} & 0.0 & \textbf{0.551} & -0.9 & \textbf{0.551} & -0.6 \\
 & Gender & 0.539 & 0.704 & -0.5 & 0.580 & -2.7 & \textbf{0.539} & +1.7 \\
 & Safety & 0.510 & 0.735 & +1.0 & 0.609 & +1.0 & \textbf{0.510} & 0.0 \\
\addlinespace[2pt]
Gemma-3-12B & Sycophancy & 0.556 & 0.613 & 0.0 & \textbf{0.556} & -0.5 & \textbf{0.556} & -0.7 \\
 & Gender & 0.539 & \textbf{0.539} & +3.5 & 0.606 & -8.7 & 0.539 & -1.0 \\
 & Safety & 0.510 & 0.589 & +3.8 & \textbf{0.510} & +3.8 & \textbf{0.510} & 0.0 \\
\addlinespace[2pt]
Gemma-3-27B & Sycophancy & 0.526 & 0.586 & -0.2 & 0.566 & -0.1 & \textbf{0.526} & +0.1 \\
 & Gender & 0.539 & 0.700 & +1.7 & 0.678 & -2.6 & \textbf{0.539} & -1.1 \\
 & Safety & 0.510 & 0.686 & +0.9 & \textbf{0.510} & +0.7 & \textbf{0.510} & 0.0 \\
\addlinespace[2pt]
\midrule
\textbf{Mean sycophancy} &  & 0.560 & 0.693 & 0.0 & 0.563   & -0.9 & \textbf{0.560} & -0.4 \\
\textbf{Mean gender} &  & 0.539 & 0.743 & +0.2 & 0.687   & -1.6 & \textbf{0.539} & -1.4 \\
\textbf{Mean safety} &  & 0.510 & 0.732 & +0.8 & 0.625   & +0.8 & \textbf{0.512} & 0.0 \\
\bottomrule
\end{tabular}
\end{table}

\begin{figure}[h]
\centering
\includegraphics[width=\linewidth]{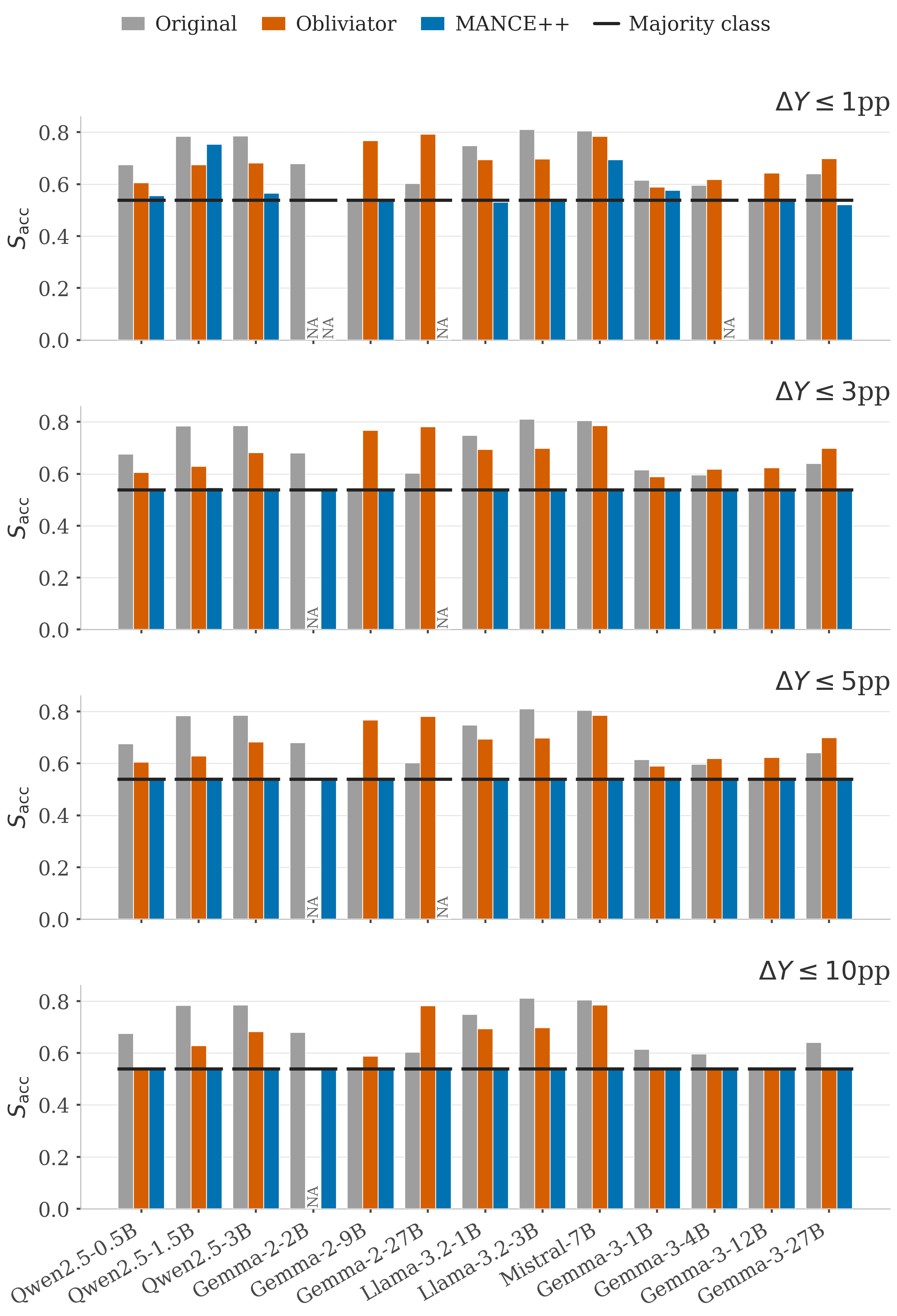}
\caption{\textbf{Gender: per-model nonlinear-probe accuracy at all four surgicality budgets.} Each row is one budget $D_Y \in \{1, 3, 5, 10\}$pp.}
\label{fig:per_model_bars_gender_full}
\end{figure}

\begin{figure}[h]
\centering
\includegraphics[width=\linewidth]{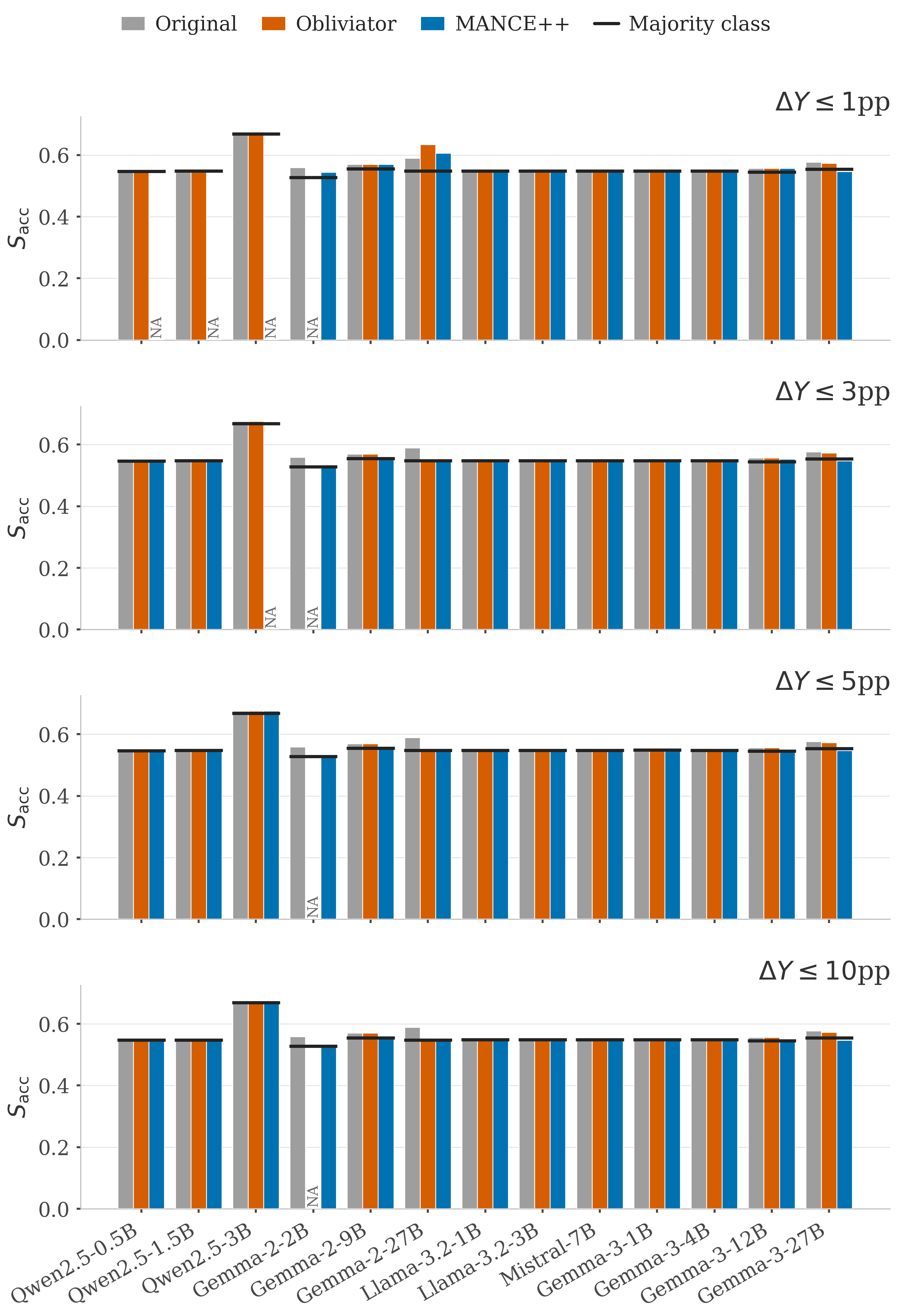}
\caption{\textbf{Sycophancy: per-model nonlinear-probe accuracy at four surgicality budgets.} Layout matches Fig.~\ref{fig:per_model_bars_gender_full}.}
\label{fig:per_model_bars_sycophancy_app}
\end{figure}

\begin{figure}[h]
\centering
\includegraphics[width=\linewidth]{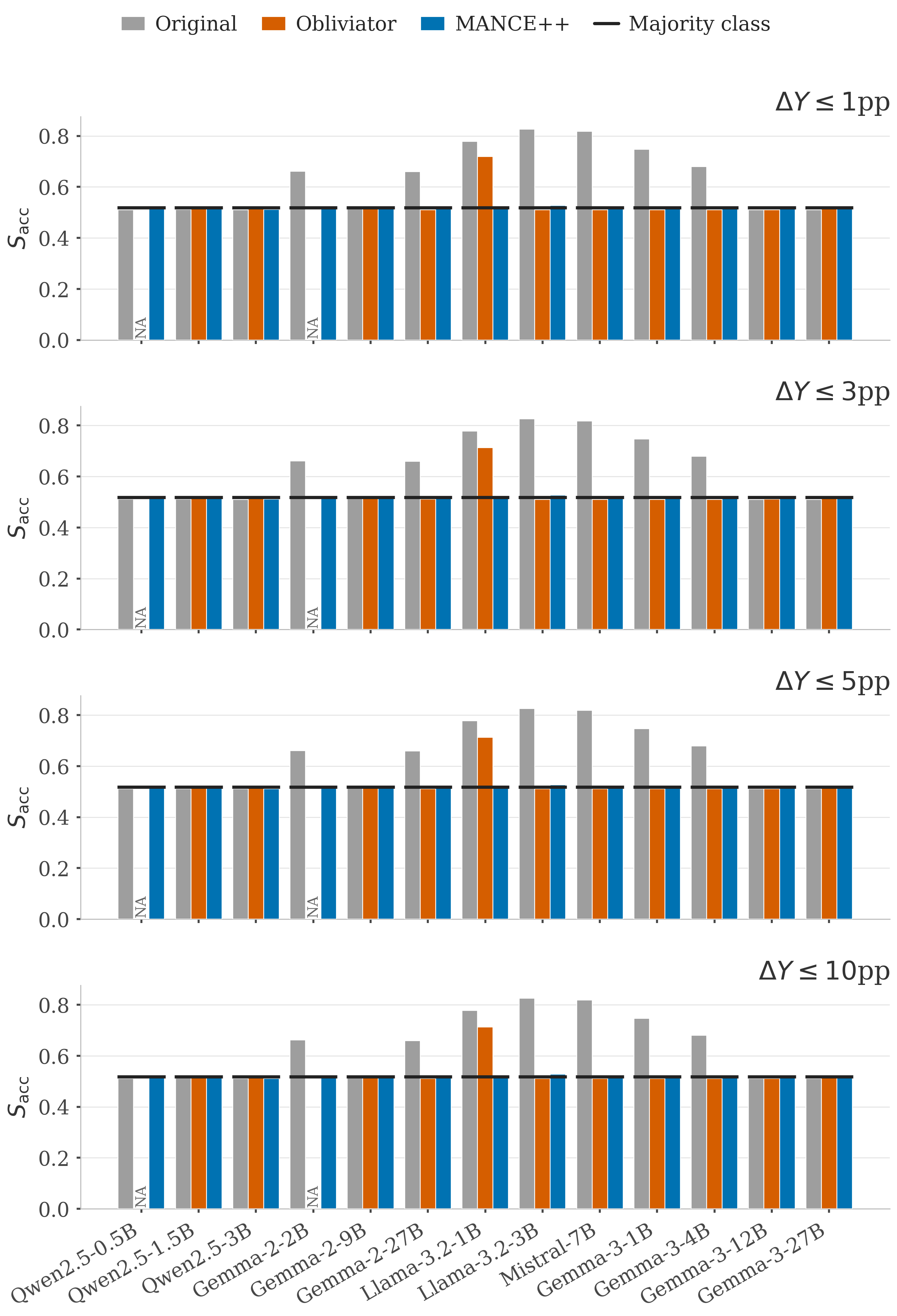}
\caption{\textbf{Safety: per-model nonlinear-probe accuracy at four surgicality budgets.} Layout matches Fig.~\ref{fig:per_model_bars_gender_full}.}
\label{fig:per_model_bars_safety_app}
\end{figure}

\section{NLP full per-setting breakdown}
\label{sec:appendix_per_panel}

Table~\ref{tab:per_panel_breakdown} reports the per-setting version of the main-text Tab.~\ref{tab:main_results_unified}, restricted to the 7 method columns shared with the main-text breakdown (the pure \MANCE{} and \MANCEp{} ablation rows of the main-text table are not repeated per-setting). For each of the 39 NLP settings and each of the four surgicality budgets, the signed deviation $D_S$ in pp is measured at the in-budget trajectory step with smallest $|D_S|$. Bold marks the row-best per setting per budget block (every method within numeric precision of the row-best is bolded). The trailing row reports the per-method mean across the 39 settings.

\begin{table}[t]
\centering
\caption[NLP per-panel leakage breakdown]{\textbf{Per-panel breakdown.} Signed $D_S$ in pp at the in-budget trajectory step with smallest $|D_S|$, subject to the surgicality budget named by each $D_Y \in \{1,3,5,10\}$pp block. Bold = best (smallest absolute deviation from floor) per panel per budget block. `--' = no in-budget step at that budget for that method. Final row: per-budget per-method mean across the 39 panels.}
\label{tab:per_panel_breakdown}
\scriptsize
\setlength{\tabcolsep}{2.2pt}
\resizebox{\textwidth}{!}{%
\begin{tabular}{l|ccccccc|ccccccc|ccccccc|ccccccc}
\toprule
Panel & \multicolumn{7}{c}{$D_Y \le 1$pp} & \multicolumn{7}{c}{$D_Y \le 3$pp} & \multicolumn{7}{c}{$D_Y \le 5$pp} & \multicolumn{7}{c}{$D_Y \le 10$pp} \\
 & LE & L+C & INLP & IGBP & Obl & Amb++ & MANCE++ & LE & L+C & INLP & IGBP & Obl & Amb++ & MANCE++ & LE & L+C & INLP & IGBP & Obl & Amb++ & MANCE++ & LE & L+C & INLP & IGBP & Obl & Amb++ & MANCE++ \\
\midrule
Gemma2-27B/Gen L23 & +12.3 & -- & +11.1 & +12.8 & \textbf{0.0} & -- & -- & +12.3 & +6.6 & +11.1 & +12.8 & \textbf{0.0} & -- & \textbf{0.0} & +12.3 & +6.6 & +11.1 & +12.8 & \textbf{0.0} & -- & \textbf{0.0} & +12.3 & +6.6 & +11.1 & +12.8 & \textbf{0.0} & -- & \textbf{0.0} \\
Gemma2-27B/Saf L23 & +22.5 & +13.3 & +28.6 & +22.3 & -0.7 & -1.7 & \textbf{0.0} & +22.5 & +13.3 & +28.6 & +22.3 & -0.7 & -1.0 & \textbf{0.0} & +22.5 & +13.3 & +28.6 & +22.3 & -0.7 & -0.1 & \textbf{0.0} & +22.5 & +13.3 & +28.6 & +22.3 & -0.7 & -0.1 & \textbf{0.0} \\
Gemma2-27B/Syc L23 & +14.3 & 0.0 & +10.7 & +14.8 & +8.5 & +18.7 & \textbf{0.0} & +14.3 & 0.0 & +10.7 & +14.8 & +0.5 & +18.7 & \textbf{0.0} & +14.3 & 0.0 & +10.7 & +14.8 & +0.5 & +15.5 & \textbf{0.0} & +14.3 & 0.0 & +10.7 & +14.8 & +0.5 & +15.5 & \textbf{0.0} \\
Gemma2-2B/Gen L13 & +18.3 & +14.2 & +15.1 & \textbf{+5.6} & +24.0 & +18.5 & -- & +18.3 & +14.2 & +15.1 & +5.6 & +21.8 & +18.5 & \textbf{0.0} & +18.3 & +14.2 & +15.1 & +5.6 & +21.8 & +18.5 & \textbf{0.0} & +18.3 & +14.2 & +15.1 & +5.6 & +21.8 & +18.5 & \textbf{0.0} \\
Gemma2-2B/Saf L13 & +20.5 & +14.3 & +12.6 & +6.7 & -0.6 & \textbf{+0.1} & -0.1 & +20.5 & +14.3 & +12.6 & +6.7 & -0.6 & \textbf{+0.1} & -0.1 & +20.5 & +14.3 & +12.6 & +6.7 & -0.6 & \textbf{+0.1} & -0.1 & +20.5 & +14.3 & +12.6 & +6.7 & -0.6 & \textbf{0.0} & -0.1 \\
Gemma2-2B/Syc L13 & +9.9 & \textbf{0.0} & +2.5 & +2.8 & +1.6 & +13.0 & -0.2 & +9.9 & \textbf{0.0} & +2.5 & +2.8 & +1.6 & +13.0 & -0.2 & +9.9 & \textbf{0.0} & +2.5 & +2.8 & +1.6 & +13.0 & -0.2 & +9.9 & \textbf{0.0} & +2.5 & +2.8 & +1.6 & +12.9 & -0.2 \\
Gemma2-9B/Gen L21 & +9.9 & +2.3 & +5.1 & +18.6 & +17.4 & +15.4 & \textbf{0.0} & +9.9 & +2.3 & +5.1 & +13.7 & +17.4 & +15.4 & \textbf{0.0} & +9.9 & +2.3 & +5.1 & +13.7 & +17.4 & +15.4 & \textbf{0.0} & +9.9 & +2.3 & +5.1 & +13.7 & +4.8 & +15.4 & \textbf{0.0} \\
Gemma2-9B/Saf L21 & +11.3 & \textbf{0.0} & +16.5 & +7.2 & -0.4 & +2.0 & 0.0 & +11.3 & \textbf{0.0} & +16.5 & +7.2 & -0.4 & +0.1 & 0.0 & +11.3 & \textbf{0.0} & +16.5 & +7.2 & -0.4 & +0.1 & 0.0 & +11.3 & \textbf{0.0} & +16.5 & +7.2 & -0.4 & +0.1 & 0.0 \\
Gemma2-9B/Syc L21 & +18.2 & \textbf{0.0} & +7.5 & +10.6 & +1.4 & +22.1 & +1.4 & +18.2 & \textbf{0.0} & +7.5 & +10.6 & +1.4 & +22.1 & +0.1 & +18.2 & \textbf{0.0} & +7.5 & +10.6 & +1.4 & +22.1 & +0.1 & +18.2 & \textbf{0.0} & +7.5 & +10.6 & +1.4 & +20.9 & +0.1 \\
Gemma3-12B/Gen L24 & 0.0 & -- & -- & -- & \textbf{0.0} & +8.8 & \textbf{0.0} & 0.0 & -- & +11.0 & -- & \textbf{0.0} & -0.9 & \textbf{0.0} & 0.0 & -- & +10.3 & -- & \textbf{0.0} & \textbf{0.0} & \textbf{0.0} & 0.0 & +6.7 & +6.3 & -- & \textbf{0.0} & \textbf{0.0} & \textbf{0.0} \\
Gemma3-12B/Saf L24 & +7.9 & \textbf{0.0} & +15.4 & +24.5 & -0.7 & +0.1 & +0.1 & +7.9 & \textbf{0.0} & +15.4 & +24.5 & -0.7 & +0.1 & +0.1 & +7.9 & \textbf{0.0} & +15.4 & +24.5 & -0.7 & +0.1 & +0.1 & +7.9 & \textbf{0.0} & +15.4 & +24.5 & -0.7 & +0.1 & +0.1 \\
Gemma3-12B/Syc L24 & +5.7 & \textbf{0.0} & +0.8 & +4.4 & +1.1 & -- & +1.1 & +5.7 & \textbf{0.0} & +0.8 & +4.4 & +1.1 & +7.7 & +0.8 & +5.7 & \textbf{0.0} & +0.8 & +4.4 & +1.1 & +7.7 & +0.2 & +5.7 & \textbf{0.0} & +0.8 & +4.4 & +1.1 & +7.7 & -0.1 \\
Gemma3-1B/Gen L13 & +16.9 & +7.4 & +8.5 & -- & +5.0 & -- & \textbf{0.0} & +16.9 & +7.4 & +2.1 & +15.0 & +5.0 & +15.4 & \textbf{0.0} & +16.9 & +7.4 & \textbf{0.0} & +12.7 & +5.0 & +15.4 & \textbf{0.0} & +16.9 & +7.4 & \textbf{0.0} & +12.7 & \textbf{0.0} & +15.4 & \textbf{0.0} \\
Gemma3-1B/Saf L13 & +21.5 & +21.2 & +16.2 & +20.6 & -0.8 & -0.4 & \textbf{0.0} & +21.5 & +21.2 & +16.2 & +20.6 & -0.8 & -0.4 & \textbf{0.0} & +21.5 & +21.2 & +16.2 & +20.6 & -0.8 & -0.4 & \textbf{0.0} & +21.5 & +21.2 & +16.2 & +20.6 & -0.8 & -0.3 & \textbf{0.0} \\
Gemma3-1B/Syc L13 & +17.6 & -- & +1.0 & +4.9 & +0.3 & +0.3 & \textbf{+0.1} & +17.6 & \textbf{0.0} & +1.0 & +4.9 & +0.3 & +0.3 & +0.1 & +17.6 & \textbf{0.0} & +1.0 & +4.9 & +0.3 & +0.3 & +0.1 & +17.6 & \textbf{0.0} & +1.0 & +4.9 & +0.3 & +0.3 & +0.1 \\
Gemma3-27B/Gen L31 & +16.0 & -- & +25.6 & +23.9 & +15.9 & -- & \textbf{0.0} & +16.0 & +13.9 & +25.6 & +23.9 & +15.9 & -- & \textbf{0.0} & +16.0 & +13.9 & +25.6 & +23.9 & +15.9 & -- & \textbf{0.0} & +16.0 & +13.9 & +24.4 & +23.9 & \textbf{0.0} & -- & \textbf{0.0} \\
Gemma3-27B/Saf L31 & +17.6 & \textbf{0.0} & +23.8 & +25.3 & -0.3 & 0.0 & -0.1 & +17.6 & \textbf{0.0} & +23.8 & +25.3 & -0.3 & 0.0 & -0.1 & +17.6 & \textbf{0.0} & +23.8 & +25.3 & -0.3 & 0.0 & -0.1 & +17.6 & \textbf{0.0} & +23.8 & +25.3 & -0.3 & 0.0 & -0.1 \\
Gemma3-27B/Syc L31 & +6.0 & +4.0 & \textbf{+0.3} & -0.3 & +1.8 & +4.3 & -0.8 & +6.0 & +4.0 & \textbf{+0.3} & -0.3 & +1.8 & +4.3 & -0.8 & +6.0 & +4.0 & \textbf{+0.3} & -0.3 & +1.8 & +4.3 & -0.8 & +6.0 & +4.0 & \textbf{+0.3} & -0.3 & +1.8 & +3.5 & -0.8 \\
Gemma3-4B/Gen L17 & +16.5 & -- & +8.5 & +19.5 & +7.9 & -- & \textbf{0.0} & +16.5 & +4.0 & +8.5 & +19.5 & +7.9 & +15.0 & \textbf{0.0} & +16.5 & +4.0 & +8.5 & +19.5 & +7.9 & +15.0 & \textbf{0.0} & +16.5 & +4.0 & +8.5 & +19.5 & \textbf{0.0} & +15.0 & \textbf{0.0} \\
Gemma3-4B/Saf L17 & +22.5 & +9.9 & +27.3 & +26.1 & -0.8 & \textbf{-0.1} & +0.1 & +22.5 & +9.9 & +27.3 & +26.1 & -0.8 & \textbf{-0.1} & +0.1 & +22.5 & +9.9 & +27.3 & +26.1 & -0.8 & \textbf{-0.1} & +0.1 & +22.5 & +9.9 & +27.3 & +26.1 & -0.8 & \textbf{-0.1} & +0.1 \\
Gemma3-4B/Syc L17 & \textbf{0.0} & \textbf{0.0} & +0.3 & +13.2 & +0.2 & -- & +0.3 & \textbf{0.0} & \textbf{0.0} & +0.3 & +13.2 & +0.2 & +0.4 & -0.1 & \textbf{0.0} & \textbf{0.0} & +0.3 & +13.2 & +0.2 & +0.4 & -0.1 & \textbf{0.0} & \textbf{0.0} & +0.3 & +13.2 & +0.2 & +0.4 & -0.1 \\
Llama3.2-1B/Gen L8 & +31.6 & +19.9 & +21.5 & +4.6 & +15.4 & -- & \textbf{+3.5} & +31.6 & +19.9 & +21.5 & +2.3 & +15.4 & -- & \textbf{+0.3} & +31.6 & +19.9 & +21.5 & +2.3 & +15.4 & +25.7 & \textbf{+0.1} & +31.6 & +19.9 & +21.5 & +2.3 & +15.4 & +25.7 & \textbf{+0.1} \\
Llama3.2-1B/Saf L8 & +30.7 & +26.7 & +28.3 & +8.8 & -0.7 & -5.7 & \textbf{-0.1} & +30.7 & +26.7 & +28.3 & +8.8 & -0.7 & -5.7 & \textbf{-0.1} & +30.7 & +26.7 & +28.3 & +8.8 & -0.7 & -5.7 & \textbf{-0.1} & +30.7 & +26.7 & +28.3 & +8.8 & -0.7 & -1.7 & \textbf{-0.1} \\
Llama3.2-1B/Syc L8 & +29.8 & \textbf{0.0} & +0.3 & +0.3 & +0.3 & -- & +0.3 & +29.8 & \textbf{0.0} & +0.3 & +0.3 & +0.3 & -- & +0.3 & +29.8 & \textbf{0.0} & +0.3 & +0.3 & +0.3 & -- & +0.3 & +29.8 & \textbf{0.0} & +0.3 & +0.3 & +0.3 & -- & +0.3 \\
Llama3.2-3B/Gen L14 & +35.0 & +25.5 & +31.7 & +2.5 & +15.8 & -- & \textbf{0.0} & +35.0 & +25.5 & +31.7 & +2.1 & +15.8 & +30.3 & \textbf{0.0} & +35.0 & +25.5 & +31.7 & +2.1 & +15.8 & +30.3 & \textbf{0.0} & +35.0 & +25.5 & +31.7 & +2.1 & +15.8 & +30.3 & \textbf{0.0} \\
Llama3.2-3B/Saf L14 & +32.7 & +31.0 & +33.1 & +9.4 & -0.8 & \textbf{-0.7} & +0.9 & +32.7 & +31.0 & +32.1 & +9.4 & -0.8 & \textbf{-0.7} & +0.9 & +32.7 & +31.0 & +32.1 & +9.4 & -0.8 & \textbf{-0.7} & +0.9 & +32.7 & +31.0 & +32.1 & +9.4 & -0.8 & \textbf{-0.7} & +0.9 \\
Llama3.2-3B/Syc L14 & +34.5 & \textbf{0.0} & +0.3 & +2.8 & +0.3 & -- & +0.2 & +34.5 & \textbf{0.0} & +0.3 & +2.8 & +0.3 & -- & +0.2 & +34.5 & \textbf{0.0} & +0.3 & +2.8 & +0.3 & -- & +0.2 & +34.5 & \textbf{0.0} & +0.3 & +2.8 & +0.3 & +0.3 & +0.2 \\
Mistral-7B/Gen L16 & +31.2 & +27.6 & +29.5 & \textbf{+8.1} & +24.6 & -- & -- & +31.2 & +27.6 & +29.5 & +8.1 & +24.6 & +28.6 & \textbf{0.0} & +31.2 & +27.6 & +29.5 & +8.1 & +24.6 & +28.6 & \textbf{0.0} & +31.2 & +27.6 & +29.5 & +8.1 & +24.6 & +28.6 & \textbf{0.0} \\
Mistral-7B/Saf L16 & +31.5 & +30.0 & +34.6 & +8.0 & -0.8 & -2.9 & \textbf{+0.1} & +31.5 & +30.0 & +34.3 & +8.0 & -0.8 & -2.9 & \textbf{+0.1} & +31.5 & +30.0 & +34.3 & +8.0 & -0.8 & -2.9 & \textbf{+0.1} & +31.5 & +30.0 & +34.3 & +8.0 & -0.8 & -2.9 & \textbf{+0.1} \\
Mistral-7B/Syc L16 & +37.0 & \textbf{0.0} & +0.3 & +0.3 & +0.3 & +36.9 & +0.3 & +37.0 & \textbf{0.0} & +0.3 & +0.3 & +0.3 & +36.7 & +0.3 & +37.0 & \textbf{0.0} & +0.3 & +0.3 & +0.3 & +36.7 & +0.3 & +37.0 & \textbf{0.0} & +0.3 & +0.3 & +0.3 & +36.7 & +0.3 \\
Qwen2.5-0.5B/Gen L12 & +17.0 & +14.5 & +13.1 & \textbf{+5.4} & +6.6 & +15.8 & +7.6 & +17.0 & +14.5 & +13.1 & +5.4 & +6.6 & +15.8 & \textbf{-0.1} & +17.0 & +14.5 & +13.1 & +5.4 & +6.6 & +15.8 & \textbf{-0.1} & +17.0 & +14.5 & +13.1 & +5.4 & \textbf{0.0} & +15.8 & \textbf{0.0} \\
Qwen2.5-0.5B/Saf L12 & +24.1 & +1.9 & +16.7 & +3.2 & -0.8 & +0.1 & \textbf{0.0} & +24.1 & +1.9 & +16.7 & +3.2 & -0.8 & +0.1 & \textbf{0.0} & +24.1 & +1.9 & +16.7 & +3.2 & -0.8 & +0.1 & \textbf{0.0} & +24.1 & +1.9 & +16.7 & +3.2 & -0.8 & +0.1 & \textbf{0.0} \\
Qwen2.5-0.5B/Syc L12 & \textbf{0.0} & -- & +0.3 & +3.3 & +0.3 & -- & -0.1 & \textbf{0.0} & \textbf{0.0} & +0.3 & +3.3 & +0.3 & -- & -0.1 & \textbf{0.0} & \textbf{0.0} & +0.3 & +3.3 & +0.3 & -- & -0.1 & \textbf{0.0} & \textbf{0.0} & +0.3 & +3.3 & +0.3 & -- & -0.1 \\
Qwen2.5-1.5B/Gen L14 & +28.2 & -- & +22.3 & \textbf{+4.1} & +13.5 & +28.2 & +21.5 & +28.2 & +24.9 & +22.3 & +4.1 & +8.9 & +28.2 & \textbf{+0.8} & +28.2 & +24.9 & +22.3 & +4.1 & +8.9 & +28.2 & \textbf{0.0} & +28.2 & +24.9 & +22.3 & +4.1 & +8.9 & +28.2 & \textbf{0.0} \\
Qwen2.5-1.5B/Saf L14 & +24.9 & \textbf{0.0} & +17.9 & +6.0 & -0.3 & +2.7 & 0.0 & +24.9 & \textbf{0.0} & +17.9 & +6.0 & -0.3 & +2.0 & 0.0 & +24.9 & \textbf{0.0} & +17.9 & +6.0 & -0.3 & +2.0 & 0.0 & +24.9 & \textbf{0.0} & +17.9 & +6.0 & -0.3 & +2.0 & 0.0 \\
Qwen2.5-1.5B/Syc L14 & \textbf{0.0} & -- & 0.0 & -0.4 & +0.3 & -- & -0.1 & \textbf{0.0} & \textbf{0.0} & 0.0 & -0.4 & +0.3 & -- & -0.1 & \textbf{0.0} & \textbf{0.0} & 0.0 & -0.4 & +0.3 & -- & -0.1 & \textbf{0.0} & \textbf{0.0} & 0.0 & -0.4 & +0.3 & -- & -0.1 \\
Qwen2.5-3B/Gen L18 & -- & +24.4 & +24.8 & \textbf{+4.7} & +14.3 & -- & +13.5 & +31.8 & +24.4 & +24.8 & +4.7 & +14.3 & +26.5 & \textbf{+0.1} & +31.8 & +24.4 & +24.8 & +4.7 & +14.3 & +26.5 & \textbf{+0.1} & +31.8 & +24.4 & +24.8 & +4.7 & +14.3 & +26.5 & \textbf{+0.1} \\
Qwen2.5-3B/Saf L18 & +20.8 & +0.9 & +22.3 & +15.4 & -0.3 & \textbf{-0.2} & -0.7 & +20.8 & +0.9 & +22.3 & +15.4 & -0.3 & \textbf{0.0} & -0.7 & +20.8 & +0.9 & +22.3 & +15.4 & -0.3 & \textbf{0.0} & -0.7 & +20.8 & +0.9 & +22.3 & +15.4 & -0.3 & \textbf{0.0} & -0.7 \\
Qwen2.5-3B/Syc L18 & \textbf{0.0} & -- & +0.6 & +0.6 & +0.6 & -- & +0.6 & \textbf{0.0} & \textbf{0.0} & +0.6 & +0.6 & +0.6 & -- & +0.6 & \textbf{0.0} & \textbf{0.0} & +0.6 & +0.6 & +0.6 & -- & +0.6 & \textbf{0.0} & \textbf{0.0} & +0.6 & +0.6 & +0.6 & -- & +0.6 \\
\midrule
Mean (pp) & +18.3 & +9.6 & +14.1 & +9.5 & +4.3 & +7.3 & +1.4 & +18.6 & +8.9 & +13.8 & +9.4 & +4.0 & +9.3 & +0.1 & +18.6 & +8.9 & +13.7 & +9.4 & +4.0 & +9.7 & 0.0 & +18.6 & +8.9 & +13.6 & +9.4 & +2.7 & +9.5 & 0.0 \\
\bottomrule
\end{tabular}%
}
\end{table}

\section{CelebA full per-attribute breakdown}
\label{sec:appendix_celeba_extended}

Tab.~\ref{tab:celeba_unified_summary} is the full per-method $\times$ per-budget summary across both surgicality regimes summarized in §\ref{sec:celeba}. Cells report signed mean $D_S$ in pp at the in-budget trajectory step with smallest $|D_S|$, plus coverage (\S\ref{sec:evaluation}) and at-chance count.

\begin{table}[t]
\centering
\caption[CelebA cross-modal results]{\textbf{CelebA cross-modal results (CLIP ViT-B/32, 40 binary attributes).} \emph{Leakage} ($\downarrow$): how much target information survives, measured as $D_S$ in pp at the within-budget step closest to chance, so $0$ is at chance and larger values mean more leakage. \emph{At Chance} ($\uparrow$), written $n/d$: of the $d$ settings a method keeps within budget (its \emph{coverage}; \S\ref{sec:evaluation}), the number $n$ it drives to chance ($|D_S| \le 0.5$pp). Each method is run on 40 attributes under each surgicality regime (5 least- vs.\ 5 most-correlated controls per target). The MANCE family is shown in three increments under one fixed configuration (eps=0.1, scorer $h{=}512$, 800 steps): \MANCE{} (manifold loop only), \MANCEp{} (LEACE $+$ manifold loop), \MANCEpp{} (LEACE $+$ rank-2 CovMatch $+$ manifold loop, the full method). Best per column in \textbf{bold}. `--' = no in-budget step at that budget for that method.}
\label{tab:celeba_unified_summary}
\footnotesize
\setlength{\tabcolsep}{4pt}
\begin{tabular}{l|cc|cc|cc|cc}
\toprule
& \multicolumn{2}{c|}{$D_Y \le 1$pp} & \multicolumn{2}{c|}{$D_Y \le 3$pp} & \multicolumn{2}{c|}{$D_Y \le 5$pp} & \multicolumn{2}{c}{$D_Y \le 10$pp} \\
Method & Leakage$\downarrow$ & Chance$\uparrow$ & Leakage$\downarrow$ & Chance$\uparrow$ & Leakage$\downarrow$ & Chance$\uparrow$ & Leakage$\downarrow$ & Chance$\uparrow$ \\
\midrule
\multicolumn{9}{l}{\textit{5 least-correlated controls (40 attributes).}} \\
\midrule
LEACE & $+6.6$ & $15/39$ & $+6.4$ & $16/40$ & $+6.4$ & $16/40$ & $+6.4$ & $16/40$ \\
LEACE$+$CovMatch & $+5.4$ & $21/37$ & $+5.1$ & $22/40$ & $+5.1$ & $22/40$ & $+5.1$ & $22/40$ \\
INLP & $+5.5$ & $26/40$ & $+5.5$ & $26/40$ & $+5.5$ & $26/40$ & $+5.5$ & $26/40$ \\
IGBP & $+5.8$ & $10/36$ & $+5.3$ & $13/40$ & $+5.3$ & $13/40$ & $+5.3$ & $13/40$ \\
Obliviator & $\mathbf{0.0}$ & $15/15$ & $\mathbf{0.0}$ & $29/29$ & $-0.4$ & $37/39$ & $-0.4$ & $38/40$ \\
\AmbCEpp{} & $+1.4$ & $15/20$ & $+2.9$ & $28/39$ & $+0.7$ & $32/40$ & $+0.4$ & $36/40$ \\
\MANCE{} & $+1.2$ & $32/40$ & $+1.3$ & $35/40$ & $+1.3$ & $35/40$ & $+1.3$ & $35/40$ \\
\MANCEp{} & $+0.4$ & $34/40$ & $\mathbf{0.0}$ & $39/40$ & $\mathbf{0.0}$ & $39/40$ & $\mathbf{0.0}$ & $39/40$ \\
\MANCEpp{} & $+0.7$ & $35/39$ & $\mathbf{0.0}$ & $39/40$ & $\mathbf{0.0}$ & $40/40$ & $\mathbf{0.0}$ & $40/40$ \\
\midrule
\multicolumn{9}{l}{\textit{5 most-correlated controls (40 attributes).}} \\
\midrule
LEACE & $+3.9$ & $11/20$ & $+6.2$ & $16/37$ & $+6.4$ & $16/40$ & $+6.4$ & $16/40$ \\
LEACE$+$CovMatch & $+0.5$ & $9/12$ & $+4.0$ & $21/34$ & $+4.7$ & $22/39$ & $+5.1$ & $22/40$ \\
INLP & $+5.8$ & $26/40$ & $+5.6$ & $26/40$ & $+5.5$ & $26/40$ & $+5.5$ & $26/40$ \\
IGBP & $+7.0$ & $13/40$ & $+5.6$ & $13/40$ & $+5.3$ & $13/40$ & $+5.3$ & $13/40$ \\
Obliviator & $\mathbf{0.0}$ & $2/2$ & $\mathbf{0.0}$ & $15/15$ & $\mathbf{0.0}$ & $30/30$ & $\mathbf{0.0}$ & $34/34$ \\
\AmbCEpp{} & $+2.2$ & $1/3$ & $+7.4$ & $10/20$ & $+5.8$ & $17/33$ & $+4.7$ & $23/40$ \\
\MANCE{} & $+4.3$ & $23/40$ & $+2.9$ & $30/40$ & $+2.1$ & $31/40$ & $+1.0$ & $31/40$ \\
\MANCEp{} & $+2.2$ & $17/22$ & $+3.1$ & $27/36$ & $+2.9$ & $31/40$ & $+1.1$ & $34/40$ \\
\MANCEpp{} & $+0.5$ & $15/19$ & $+2.6$ & $28/34$ & $+2.5$ & $33/38$ & $+1.7$ & $36/40$ \\

\bottomrule
\end{tabular}
\end{table}

\begin{figure}[h]
\centering
\includegraphics[width=\linewidth]{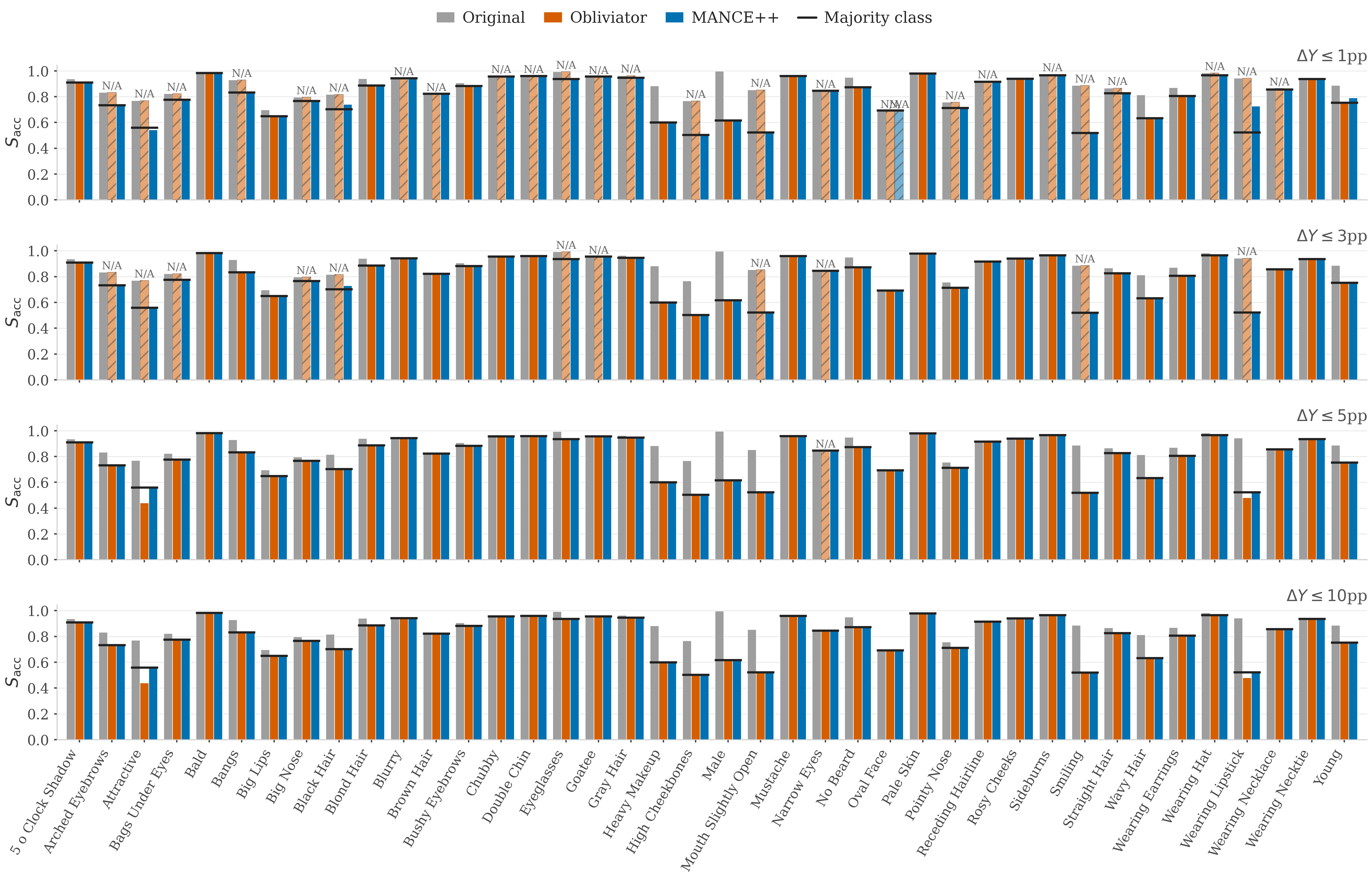}
\caption*{(a) 5 least-correlated controls (highly disentangled control concepts).}
\vspace{0.5em}
\includegraphics[width=\linewidth]{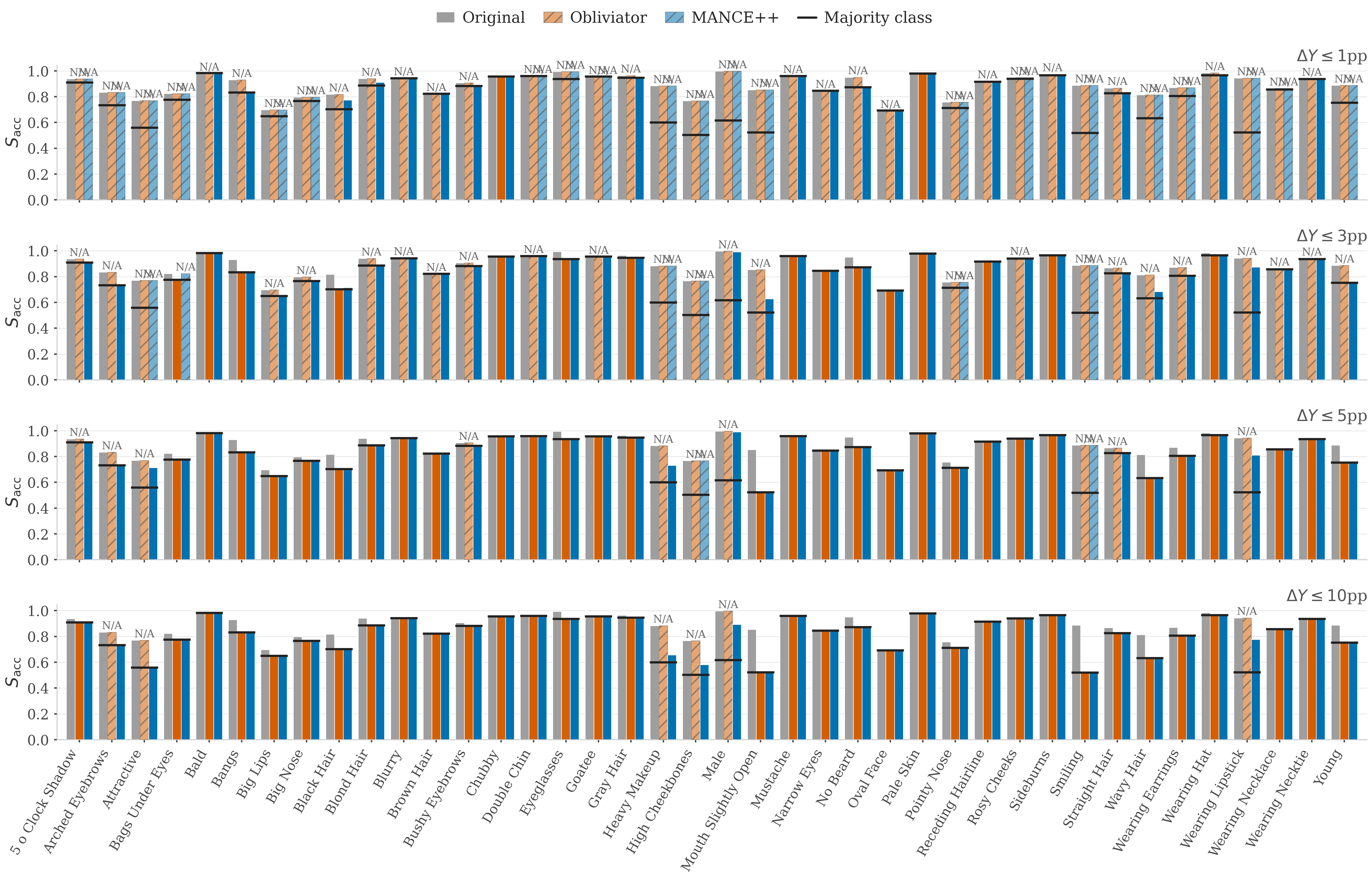}
\caption*{(b) 5 most-correlated controls (highly entangled control concepts).}
\caption{\textbf{CelebA per-attribute nonlinear-probe accuracy at four surgicality budgets.} Same bar-chart layout as the NLP per-model figures. For each attribute, three bars: Clean (gray, no erasure), \MANCEpp{} (blue, the proposed method), Obliviator (orange, baseline). Black tick: per-attribute majority-class floor (lower bars are better; bars at the floor mean the probe is at chance). Hatched bars marked ``N/A'' = the method had no in-budget step at that $D_Y$; we substitute its clean (un-erased) $S$ as a stand-in, so a hatched-orange tall bar next to a low solid-blue bar is a clean \MANCEpp{} win on a setting Obliviator could not reach. Most-correlated regime at $D_Y \le 1$pp: Obliviator is marked N/A on $38/40$ attributes, while \MANCEpp{} reaches the budget on $19/40$ and drives $15$ of those to chance.}
\label{fig:per_attr_bars_celeba}
\end{figure}

Fig.~\ref{fig:per_attr_bars_celeba} reports the per-attribute nonlinear-probe accuracy $S$ for each of the 40 CelebA attributes at all four surgicality budgets, in the same bar-chart style as the NLP per-model figures (App.~Figs.~\ref{fig:per_model_bars_gender_full}--\ref{fig:per_model_bars_safety_app}): three bars per attribute (Clean / \MANCEpp{} / Obliviator) and a horizontal tick at the per-attribute majority-class floor. Hatched bars annotated ``N/A'' indicate that the method had no in-budget step at the shown $D_Y$ budget; we substitute its clean (un-erased) $S$ so the visual height communicates ``no valid in-budget point'' rather than a real measurement. A clear win for \MANCEpp{} reads as a low solid blue bar paired with a high (often hatched-N/A) orange Obliviator bar, common at $D_Y \le 1$pp in the most-correlated regime, where Obliviator overshoots the budget on $38$ of $40$ high-correlation targets while \MANCEpp{} keeps $19$ in budget and drives $15$ of them to floor.

Tabs.~\ref{tab:celeba_per_panel_breakdown_dy1}--\ref{tab:celeba_per_panel_breakdown_dy5} report the per-attribute numerical version of App.~Tab.~\ref{tab:celeba_unified_summary}, one table per surgicality budget $D_Y \in \{1, 3, 5\}$pp: signed $D_S$ in pp at the in-budget trajectory step with smallest $|D_S|$. Bold marks the row-best per regime block. The trailing row reports the per-regime per-method mean. The tightest budget ($\le 1$pp) is where the gap between \MANCEpp{} and the baselines is largest; it is also where Obliviator overshoots most often and shows ``--'' on the hardest entangled targets.

\begin{table}[t]
\centering
\caption[CelebA per-attribute breakdown at 1pp budget]{\textbf{CelebA per-attribute breakdown at $D_Y \le 1$pp (40 attributes $\times$ 2 surgicality regimes).} Signed $D_S$ in pp at the in-budget trajectory step with smallest $|D_S|$, subject to surgicality budget $D_Y \le 1$pp on the 5 controls. Bold = best (smallest absolute deviation from floor) per attribute per regime block. `--' = no in-budget step at that budget for that method. Final row: per-regime per-method mean.}
\label{tab:celeba_per_panel_breakdown_dy1}
\scriptsize
\setlength{\tabcolsep}{2.2pt}
\resizebox{\textwidth}{!}{%
\begin{tabular}{l|ccccccc|ccccccc}
\toprule
Concept & \multicolumn{7}{c}{5 least-correlated controls} & \multicolumn{7}{c}{5 most-correlated controls} \\
 & LE & L+C & INLP & IGBP & Obl & Amb++ & MANCE++ & LE & L+C & INLP & IGBP & Obl & Amb++ & MANCE++ \\
\midrule
5\_o\_Clock\_Shadow & +3.0 & +1.3 & \textbf{0.0} & +3.0 & \textbf{0.0} & +1.3 & \textbf{0.0} & -- & -- & \textbf{0.0} & +3.0 & -- & -- & -- \\
Arched\_Eyebrows & +9.7 & +3.0 & +6.7 & +6.7 & -- & +4.7 & \textbf{0.0} & -- & -- & \textbf{+6.7} & \textbf{+6.7} & -- & -- & -- \\
Attractive & +15.3 & +17.3 & +17.7 & \textbf{+1.0} & -- & -- & -1.7 & -- & -- & \textbf{+17.0} & +18.0 & -- & -- & -- \\
Bags\_Under\_Eyes & +0.3 & \textbf{0.0} & \textbf{0.0} & +0.3 & -- & -- & \textbf{0.0} & -- & -- & \textbf{0.0} & +0.3 & -- & -- & -- \\
Bald & \textbf{0.0} & \textbf{0.0} & \textbf{0.0} & -- & \textbf{0.0} & -- & \textbf{0.0} & \textbf{0.0} & \textbf{0.0} & \textbf{0.0} & \textbf{0.0} & -- & -- & \textbf{0.0} \\
Bangs & +4.7 & \textbf{0.0} & \textbf{0.0} & +3.7 & -- & -- & \textbf{0.0} & -- & -- & \textbf{0.0} & +3.7 & -- & -- & +0.3 \\
Big\_Lips & +1.7 & \textbf{0.0} & \textbf{0.0} & \textbf{0.0} & \textbf{0.0} & \textbf{0.0} & \textbf{0.0} & -- & -- & \textbf{0.0} & \textbf{0.0} & -- & -- & -- \\
Big\_Nose & +2.3 & +2.3 & \textbf{0.0} & -- & -- & -- & \textbf{0.0} & -- & -- & \textbf{0.0} & +2.0 & -- & -- & -- \\
Black\_Hair & +10.7 & -- & +8.3 & +4.0 & -- & -- & \textbf{+3.7} & +10.7 & +7.7 & +8.3 & \textbf{+4.0} & -- & -- & +7.0 \\
Blond\_Hair & +5.0 & +3.0 & +3.0 & +3.3 & \textbf{0.0} & \textbf{0.0} & \textbf{0.0} & +5.0 & -- & +3.0 & +3.3 & -- & -- & \textbf{+2.3} \\
Blurry & \textbf{0.0} & \textbf{0.0} & \textbf{0.0} & \textbf{0.0} & -- & \textbf{0.0} & \textbf{0.0} & -- & -- & \textbf{0.0} & \textbf{0.0} & -- & -- & \textbf{0.0} \\
Brown\_Hair & +0.7 & -- & \textbf{0.0} & +0.7 & -- & -- & \textbf{0.0} & +0.7 & -3.7 & \textbf{0.0} & +0.7 & -- & -- & \textbf{0.0} \\
Bushy\_Eyebrows & +1.0 & -0.7 & +0.7 & +3.7 & \textbf{0.0} & -- & \textbf{0.0} & +1.0 & -- & +0.7 & +3.7 & -- & -- & \textbf{-0.7} \\
Chubby & \textbf{0.0} & \textbf{0.0} & \textbf{0.0} & +0.3 & -- & +0.7 & \textbf{0.0} & \textbf{0.0} & \textbf{0.0} & \textbf{0.0} & +0.3 & \textbf{0.0} & -- & \textbf{0.0} \\
Double\_Chin & \textbf{0.0} & \textbf{0.0} & \textbf{0.0} & \textbf{0.0} & -- & \textbf{0.0} & \textbf{0.0} & \textbf{0.0} & -- & \textbf{0.0} & \textbf{0.0} & -- & \textbf{0.0} & -- \\
Eyeglasses & +2.7 & \textbf{0.0} & \textbf{0.0} & +6.0 & -- & -- & \textbf{0.0} & +2.7 & -- & \textbf{0.0} & +6.0 & -- & -- & -- \\
Goatee & \textbf{0.0} & \textbf{0.0} & \textbf{0.0} & -0.3 & -- & -- & \textbf{0.0} & \textbf{0.0} & -- & \textbf{0.0} & -0.3 & -- & -- & -- \\
Gray\_Hair & \textbf{0.0} & \textbf{0.0} & \textbf{0.0} & +2.0 & -- & -- & \textbf{0.0} & \textbf{0.0} & \textbf{0.0} & \textbf{0.0} & +2.0 & -- & -- & \textbf{0.0} \\
Heavy\_Makeup & +25.0 & +26.0 & +26.0 & +23.7 & \textbf{0.0} & -4.0 & \textbf{0.0} & -- & -- & \textbf{+26.3} & \textbf{+26.3} & -- & -- & -- \\
High\_Cheekbones & +21.7 & +18.7 & +21.3 & +14.7 & -- & -- & \textbf{0.0} & -- & -- & \textbf{+25.0} & +27.0 & -- & -- & -- \\
Male & +37.3 & +37.7 & +36.3 & +33.7 & \textbf{0.0} & \textbf{0.0} & \textbf{0.0} & +37.3 & -- & \textbf{+36.3} & +37.0 & -- & -- & -- \\
Mouth\_Slightly\_Open & +13.7 & +11.7 & +14.0 & +5.3 & -- & -- & \textbf{-0.3} & -- & -- & +14.0 & \textbf{+8.7} & -- & -- & -- \\
Mustache & \textbf{0.0} & \textbf{0.0} & \textbf{0.0} & \textbf{0.0} & \textbf{0.0} & \textbf{0.0} & \textbf{0.0} & \textbf{0.0} & \textbf{0.0} & \textbf{0.0} & \textbf{0.0} & -- & -- & \textbf{0.0} \\
Narrow\_Eyes & \textbf{0.0} & \textbf{0.0} & \textbf{0.0} & \textbf{0.0} & -- & -- & \textbf{0.0} & \textbf{0.0} & \textbf{0.0} & \textbf{0.0} & \textbf{0.0} & -- & -- & \textbf{0.0} \\
No\_Beard & +7.0 & +1.7 & +5.7 & +8.0 & \textbf{0.0} & \textbf{0.0} & \textbf{0.0} & +7.0 & +1.7 & +5.7 & +8.0 & -- & -- & \textbf{0.0} \\
Oval\_Face & +0.3 & -- & \textbf{0.0} & -- & -- & -- & -- & -- & -- & \textbf{0.0} & \textbf{0.0} & -- & -- & +0.7 \\
Pale\_Skin & \textbf{0.0} & \textbf{0.0} & \textbf{0.0} & \textbf{0.0} & \textbf{0.0} & -- & \textbf{0.0} & \textbf{0.0} & \textbf{0.0} & \textbf{0.0} & \textbf{0.0} & \textbf{0.0} & -- & \textbf{0.0} \\
Pointy\_Nose & -5.0 & -7.3 & -0.3 & +2.0 & -- & \textbf{0.0} & \textbf{0.0} & -- & -- & \textbf{-0.3} & +2.0 & -- & -2.0 & -- \\
Receding\_Hairline & -- & \textbf{0.0} & \textbf{0.0} & -- & -- & \textbf{0.0} & \textbf{0.0} & \textbf{0.0} & \textbf{0.0} & \textbf{0.0} & +0.3 & -- & -- & \textbf{0.0} \\
Rosy\_Cheeks & +0.3 & \textbf{0.0} & \textbf{0.0} & +1.3 & \textbf{0.0} & \textbf{0.0} & \textbf{0.0} & -- & -- & \textbf{0.0} & +1.3 & -- & -- & -- \\
Sideburns & +0.3 & \textbf{0.0} & \textbf{0.0} & +0.7 & -- & \textbf{0.0} & \textbf{0.0} & +0.3 & \textbf{0.0} & \textbf{0.0} & +0.7 & -- & -- & \textbf{0.0} \\
Smiling & +30.3 & +26.7 & +22.3 & +16.0 & -- & +25.3 & \textbf{0.0} & -- & -- & \textbf{+29.3} & +34.3 & -- & -- & -- \\
Straight\_Hair & \textbf{0.0} & +0.3 & \textbf{0.0} & +2.3 & -- & \textbf{0.0} & \textbf{0.0} & \textbf{0.0} & +0.3 & \textbf{0.0} & +2.3 & -- & -- & \textbf{0.0} \\
Wavy\_Hair & +14.0 & +10.7 & +8.0 & +15.3 & \textbf{0.0} & \textbf{0.0} & \textbf{0.0} & -- & -- & \textbf{+8.0} & +15.3 & -- & -- & -- \\
Wearing\_Earrings & +2.7 & \textbf{0.0} & +0.3 & +6.7 & \textbf{0.0} & \textbf{0.0} & \textbf{0.0} & -- & -- & \textbf{+0.3} & +6.7 & -- & -- & -- \\
Wearing\_Hat & \textbf{0.0} & \textbf{0.0} & \textbf{0.0} & +2.3 & -- & \textbf{0.0} & \textbf{0.0} & -- & -- & \textbf{0.0} & +2.3 & -- & -- & \textbf{0.0} \\
Wearing\_Lipstick & +40.0 & +38.0 & +39.0 & +27.7 & -- & -- & \textbf{+20.3} & -- & -- & +39.0 & \textbf{+38.3} & -- & -- & -- \\
Wearing\_Necklace & -1.0 & \textbf{0.0} & \textbf{0.0} & \textbf{0.0} & -- & -- & \textbf{0.0} & -- & -- & \textbf{0.0} & \textbf{0.0} & -- & -- & -- \\
Wearing\_Necktie & +1.3 & \textbf{0.0} & \textbf{0.0} & +2.0 & \textbf{0.0} & -- & \textbf{0.0} & +1.3 & -- & \textbf{0.0} & +2.0 & -- & -- & \textbf{0.0} \\
Young & +11.0 & +9.3 & +11.0 & +12.3 & \textbf{0.0} & -- & +3.7 & +11.0 & -- & +11.0 & +12.3 & -- & \textbf{+8.7} & -- \\
\midrule
Mean (pp) & +6.6 & +5.4 & +5.5 & +5.8 & 0.0 & +1.4 & +0.7 & +3.9 & +0.5 & +5.8 & +7.0 & 0.0 & +2.2 & +0.5 \\
\bottomrule
\end{tabular}%
}
\end{table}

\begin{table}[t]
\centering
\caption[CelebA per-attribute breakdown at 3pp budget]{\textbf{CelebA per-attribute breakdown at $D_Y \le 3$pp (40 attributes $\times$ 2 surgicality regimes).} Signed $D_S$ in pp at the in-budget trajectory step with smallest $|D_S|$, subject to surgicality budget $D_Y \le 3$pp on the 5 controls. Bold = best (smallest absolute deviation from floor) per attribute per regime block. `--' = no in-budget step at that budget for that method. Final row: per-regime per-method mean.}
\label{tab:celeba_per_panel_breakdown_dy3}
\scriptsize
\setlength{\tabcolsep}{2.2pt}
\resizebox{\textwidth}{!}{%
\begin{tabular}{l|ccccccc|ccccccc}
\toprule
Concept & \multicolumn{7}{c}{5 least-correlated controls} & \multicolumn{7}{c}{5 most-correlated controls} \\
 & LE & L+C & INLP & IGBP & Obl & Amb++ & MANCE++ & LE & L+C & INLP & IGBP & Obl & Amb++ & MANCE++ \\
\midrule
5\_o\_Clock\_Shadow & +3.0 & +1.3 & \textbf{0.0} & +3.0 & \textbf{0.0} & \textbf{0.0} & \textbf{0.0} & +3.0 & +1.3 & \textbf{0.0} & +3.0 & -- & -- & \textbf{0.0} \\
Arched\_Eyebrows & +9.7 & +3.0 & +6.7 & +6.7 & -- & +4.7 & \textbf{0.0} & -- & +3.0 & +6.7 & +6.7 & -- & -- & \textbf{0.0} \\
Attractive & +15.3 & +17.3 & +17.7 & +1.0 & -- & -- & \textbf{-0.3} & +15.3 & -- & +17.0 & \textbf{+1.0} & -- & -- & -- \\
Bags\_Under\_Eyes & +0.3 & \textbf{0.0} & \textbf{0.0} & +0.3 & -- & \textbf{0.0} & \textbf{0.0} & +0.3 & -- & \textbf{0.0} & +0.3 & \textbf{0.0} & -- & -- \\
Bald & \textbf{0.0} & \textbf{0.0} & \textbf{0.0} & \textbf{0.0} & \textbf{0.0} & \textbf{0.0} & \textbf{0.0} & \textbf{0.0} & \textbf{0.0} & \textbf{0.0} & \textbf{0.0} & \textbf{0.0} & \textbf{0.0} & \textbf{0.0} \\
Bangs & +4.7 & \textbf{0.0} & \textbf{0.0} & +3.7 & \textbf{0.0} & +1.7 & \textbf{0.0} & +4.7 & \textbf{0.0} & \textbf{0.0} & +3.7 & \textbf{0.0} & -- & \textbf{0.0} \\
Big\_Lips & +1.7 & \textbf{0.0} & \textbf{0.0} & \textbf{0.0} & \textbf{0.0} & \textbf{0.0} & \textbf{0.0} & +1.7 & \textbf{0.0} & \textbf{0.0} & \textbf{0.0} & -- & -- & \textbf{0.0} \\
Big\_Nose & +2.3 & +2.3 & \textbf{0.0} & +2.0 & -- & +1.3 & \textbf{0.0} & +2.3 & +2.3 & \textbf{0.0} & +2.0 & -- & +1.3 & -0.7 \\
Black\_Hair & +10.7 & +7.7 & +8.3 & +4.0 & -- & +6.7 & \textbf{+2.7} & +10.7 & +7.7 & +8.3 & +4.0 & \textbf{0.0} & +6.7 & +1.0 \\
Blond\_Hair & +5.0 & +3.0 & +3.0 & +3.3 & \textbf{0.0} & \textbf{0.0} & \textbf{0.0} & +5.0 & +3.0 & +3.0 & +3.3 & -- & +5.3 & \textbf{0.0} \\
Blurry & \textbf{0.0} & \textbf{0.0} & \textbf{0.0} & \textbf{0.0} & \textbf{0.0} & \textbf{0.0} & \textbf{0.0} & \textbf{0.0} & \textbf{0.0} & \textbf{0.0} & \textbf{0.0} & -- & \textbf{0.0} & \textbf{0.0} \\
Brown\_Hair & +0.7 & -3.7 & \textbf{0.0} & +0.7 & \textbf{0.0} & -0.3 & \textbf{0.0} & +0.7 & -3.7 & \textbf{0.0} & +0.7 & -- & -- & \textbf{0.0} \\
Bushy\_Eyebrows & +1.0 & -0.7 & +0.7 & +3.7 & \textbf{0.0} & -0.3 & \textbf{0.0} & +1.0 & -0.7 & +0.7 & +3.7 & -- & -- & \textbf{0.0} \\
Chubby & \textbf{0.0} & \textbf{0.0} & \textbf{0.0} & +0.3 & \textbf{0.0} & +0.7 & \textbf{0.0} & \textbf{0.0} & \textbf{0.0} & \textbf{0.0} & +0.3 & \textbf{0.0} & \textbf{0.0} & \textbf{0.0} \\
Double\_Chin & \textbf{0.0} & \textbf{0.0} & \textbf{0.0} & \textbf{0.0} & \textbf{0.0} & \textbf{0.0} & \textbf{0.0} & \textbf{0.0} & \textbf{0.0} & \textbf{0.0} & \textbf{0.0} & -- & \textbf{0.0} & \textbf{0.0} \\
Eyeglasses & +2.7 & \textbf{0.0} & \textbf{0.0} & +6.0 & -- & \textbf{0.0} & \textbf{0.0} & +2.7 & \textbf{0.0} & \textbf{0.0} & +6.0 & \textbf{0.0} & -- & \textbf{0.0} \\
Goatee & \textbf{0.0} & \textbf{0.0} & \textbf{0.0} & -0.3 & -- & \textbf{0.0} & \textbf{0.0} & \textbf{0.0} & \textbf{0.0} & \textbf{0.0} & -0.3 & -- & -- & \textbf{0.0} \\
Gray\_Hair & \textbf{0.0} & \textbf{0.0} & \textbf{0.0} & +2.0 & \textbf{0.0} & \textbf{0.0} & \textbf{0.0} & \textbf{0.0} & \textbf{0.0} & \textbf{0.0} & +2.0 & \textbf{0.0} & \textbf{0.0} & \textbf{0.0} \\
Heavy\_Makeup & +25.0 & +26.0 & +26.0 & +23.7 & \textbf{0.0} & \textbf{0.0} & \textbf{0.0} & +25.0 & +26.0 & +26.0 & \textbf{+23.7} & -- & +25.3 & -- \\
High\_Cheekbones & +21.7 & +18.7 & +21.3 & +14.7 & \textbf{0.0} & +17.0 & \textbf{0.0} & -- & -- & +23.3 & \textbf{+14.7} & -- & -- & -- \\
Male & +37.3 & +37.7 & +36.3 & +33.7 & \textbf{0.0} & \textbf{0.0} & \textbf{0.0} & +37.3 & +37.7 & +36.3 & \textbf{+34.0} & -- & +37.3 & +37.3 \\
Mouth\_Slightly\_Open & +13.7 & +11.7 & +14.0 & +5.3 & -- & +12.7 & \textbf{-0.3} & +13.7 & +11.7 & +14.0 & \textbf{+5.3} & -- & -- & +10.3 \\
Mustache & \textbf{0.0} & \textbf{0.0} & \textbf{0.0} & \textbf{0.0} & \textbf{0.0} & \textbf{0.0} & \textbf{0.0} & \textbf{0.0} & \textbf{0.0} & \textbf{0.0} & \textbf{0.0} & \textbf{0.0} & -- & \textbf{0.0} \\
Narrow\_Eyes & \textbf{0.0} & \textbf{0.0} & \textbf{0.0} & \textbf{0.0} & -- & \textbf{0.0} & \textbf{0.0} & \textbf{0.0} & \textbf{0.0} & \textbf{0.0} & \textbf{0.0} & \textbf{0.0} & \textbf{0.0} & \textbf{0.0} \\
No\_Beard & +7.0 & +1.7 & +5.7 & +8.0 & \textbf{0.0} & \textbf{0.0} & \textbf{0.0} & +7.0 & +1.7 & +5.7 & +8.0 & \textbf{0.0} & +2.3 & \textbf{0.0} \\
Oval\_Face & +0.3 & \textbf{0.0} & \textbf{0.0} & \textbf{0.0} & \textbf{0.0} & -4.0 & \textbf{0.0} & +0.3 & \textbf{0.0} & \textbf{0.0} & \textbf{0.0} & \textbf{0.0} & -- & \textbf{0.0} \\
Pale\_Skin & \textbf{0.0} & \textbf{0.0} & \textbf{0.0} & \textbf{0.0} & \textbf{0.0} & \textbf{0.0} & \textbf{0.0} & \textbf{0.0} & \textbf{0.0} & \textbf{0.0} & \textbf{0.0} & \textbf{0.0} & -- & \textbf{0.0} \\
Pointy\_Nose & -5.0 & -7.3 & -0.3 & +2.0 & \textbf{0.0} & \textbf{0.0} & \textbf{0.0} & -- & -- & \textbf{-0.3} & +2.0 & -- & -2.0 & -- \\
Receding\_Hairline & \textbf{0.0} & \textbf{0.0} & \textbf{0.0} & +0.3 & \textbf{0.0} & \textbf{0.0} & \textbf{0.0} & \textbf{0.0} & \textbf{0.0} & \textbf{0.0} & +0.3 & \textbf{0.0} & \textbf{0.0} & \textbf{0.0} \\
Rosy\_Cheeks & +0.3 & \textbf{0.0} & \textbf{0.0} & +1.3 & \textbf{0.0} & \textbf{0.0} & \textbf{0.0} & +0.3 & \textbf{0.0} & \textbf{0.0} & +1.3 & -- & -- & \textbf{0.0} \\
Sideburns & +0.3 & \textbf{0.0} & \textbf{0.0} & +0.7 & \textbf{0.0} & \textbf{0.0} & \textbf{0.0} & +0.3 & \textbf{0.0} & \textbf{0.0} & +0.7 & \textbf{0.0} & -0.3 & \textbf{0.0} \\
Smiling & +30.3 & +26.7 & +22.3 & +16.0 & -- & +25.3 & \textbf{0.0} & +30.3 & -- & \textbf{+24.3} & +25.0 & -- & +25.3 & -- \\
Straight\_Hair & \textbf{0.0} & +0.3 & \textbf{0.0} & +2.3 & \textbf{0.0} & \textbf{0.0} & \textbf{0.0} & \textbf{0.0} & +0.3 & \textbf{0.0} & +2.3 & -- & +0.3 & \textbf{0.0} \\
Wavy\_Hair & +14.0 & +10.7 & +8.0 & +15.3 & \textbf{0.0} & \textbf{0.0} & \textbf{0.0} & +14.0 & -- & +8.0 & +15.3 & -- & -- & \textbf{+5.0} \\
Wearing\_Earrings & +2.7 & \textbf{0.0} & +0.3 & +6.7 & \textbf{0.0} & \textbf{0.0} & \textbf{0.0} & +2.7 & \textbf{0.0} & +0.3 & +6.7 & -- & -- & \textbf{0.0} \\
Wearing\_Hat & \textbf{0.0} & \textbf{0.0} & \textbf{0.0} & +2.3 & \textbf{0.0} & \textbf{0.0} & \textbf{0.0} & \textbf{0.0} & \textbf{0.0} & \textbf{0.0} & +2.3 & \textbf{0.0} & \textbf{0.0} & \textbf{0.0} \\
Wearing\_Lipstick & +40.0 & +38.0 & +39.0 & +27.7 & -- & +38.3 & \textbf{-0.3} & +40.0 & +38.0 & +39.0 & \textbf{+31.3} & -- & +38.3 & +35.0 \\
Wearing\_Necklace & -1.0 & \textbf{0.0} & \textbf{0.0} & \textbf{0.0} & \textbf{0.0} & \textbf{0.0} & \textbf{0.0} & -1.0 & \textbf{0.0} & \textbf{0.0} & \textbf{0.0} & -- & -- & \textbf{0.0} \\
Wearing\_Necktie & +1.3 & \textbf{0.0} & \textbf{0.0} & +2.0 & \textbf{0.0} & \textbf{0.0} & \textbf{0.0} & +1.3 & \textbf{0.0} & \textbf{0.0} & +2.0 & -- & -- & \textbf{0.0} \\
Young & +11.0 & +9.3 & +11.0 & +12.3 & \textbf{0.0} & +8.7 & \textbf{0.0} & +11.0 & +9.3 & +11.0 & +12.3 & -- & +8.7 & \textbf{+0.3} \\
\midrule
Mean (pp) & +6.4 & +5.1 & +5.5 & +5.3 & 0.0 & +2.9 & 0.0 & +6.2 & +4.0 & +5.6 & +5.6 & 0.0 & +7.4 & +2.6 \\
\bottomrule
\end{tabular}%
}
\end{table}

\begin{table}[t]
\centering
\caption[CelebA per-attribute breakdown at 5pp budget]{\textbf{CelebA per-attribute breakdown at $D_Y \le 5$pp (40 attributes $\times$ 2 surgicality regimes).} Signed $D_S$ in pp at the in-budget trajectory step with smallest $|D_S|$, subject to surgicality budget $D_Y \le 5$pp on the 5 controls. Bold = best (smallest absolute deviation from floor) per attribute per regime block. `--' = no in-budget step at that budget for that method. Final row: per-regime per-method mean.}
\label{tab:celeba_per_panel_breakdown_dy5}
\scriptsize
\setlength{\tabcolsep}{2.2pt}
\resizebox{\textwidth}{!}{%
\begin{tabular}{l|ccccccc|ccccccc}
\toprule
Concept & \multicolumn{7}{c}{5 least-correlated controls} & \multicolumn{7}{c}{5 most-correlated controls} \\
 & LE & L+C & INLP & IGBP & Obl & Amb++ & MANCE++ & LE & L+C & INLP & IGBP & Obl & Amb++ & MANCE++ \\
\midrule
5\_o\_Clock\_Shadow & +3.0 & +1.3 & \textbf{0.0} & +3.0 & \textbf{0.0} & \textbf{0.0} & \textbf{0.0} & +3.0 & +1.3 & \textbf{0.0} & +3.0 & -- & +1.3 & \textbf{0.0} \\
Arched\_Eyebrows & +9.7 & +3.0 & +6.7 & +6.7 & \textbf{0.0} & -1.7 & \textbf{0.0} & +9.7 & +3.0 & +6.7 & +6.7 & -- & +4.7 & \textbf{0.0} \\
Attractive & +15.3 & +17.3 & +17.7 & +1.0 & -12.0 & +15.7 & \textbf{-0.3} & +15.3 & +17.3 & +17.0 & \textbf{+1.0} & -- & +16.0 & +15.3 \\
Bags\_Under\_Eyes & +0.3 & \textbf{0.0} & \textbf{0.0} & +0.3 & \textbf{0.0} & \textbf{0.0} & \textbf{0.0} & +0.3 & \textbf{0.0} & \textbf{0.0} & +0.3 & \textbf{0.0} & \textbf{0.0} & \textbf{0.0} \\
Bald & \textbf{0.0} & \textbf{0.0} & \textbf{0.0} & \textbf{0.0} & \textbf{0.0} & \textbf{0.0} & \textbf{0.0} & \textbf{0.0} & \textbf{0.0} & \textbf{0.0} & \textbf{0.0} & \textbf{0.0} & \textbf{0.0} & \textbf{0.0} \\
Bangs & +4.7 & \textbf{0.0} & \textbf{0.0} & +3.7 & \textbf{0.0} & +1.7 & \textbf{0.0} & +4.7 & \textbf{0.0} & \textbf{0.0} & +3.7 & \textbf{0.0} & +1.7 & \textbf{0.0} \\
Big\_Lips & +1.7 & \textbf{0.0} & \textbf{0.0} & \textbf{0.0} & \textbf{0.0} & \textbf{0.0} & \textbf{0.0} & +1.7 & \textbf{0.0} & \textbf{0.0} & \textbf{0.0} & \textbf{0.0} & +0.7 & \textbf{0.0} \\
Big\_Nose & +2.3 & +2.3 & \textbf{0.0} & +2.0 & \textbf{0.0} & \textbf{0.0} & \textbf{0.0} & +2.3 & +2.3 & \textbf{0.0} & +2.0 & \textbf{0.0} & \textbf{0.0} & \textbf{0.0} \\
Black\_Hair & +10.7 & +7.7 & +8.3 & +4.0 & \textbf{0.0} & +6.7 & +0.3 & +10.7 & +7.7 & +8.3 & +4.0 & \textbf{0.0} & \textbf{0.0} & \textbf{0.0} \\
Blond\_Hair & +5.0 & +3.0 & +3.0 & +3.3 & \textbf{0.0} & \textbf{0.0} & \textbf{0.0} & +5.0 & +3.0 & +3.0 & +3.3 & \textbf{0.0} & +5.3 & \textbf{0.0} \\
Blurry & \textbf{0.0} & \textbf{0.0} & \textbf{0.0} & \textbf{0.0} & \textbf{0.0} & \textbf{0.0} & \textbf{0.0} & \textbf{0.0} & \textbf{0.0} & \textbf{0.0} & \textbf{0.0} & \textbf{0.0} & \textbf{0.0} & \textbf{0.0} \\
Brown\_Hair & +0.7 & -3.7 & \textbf{0.0} & +0.7 & \textbf{0.0} & -0.3 & \textbf{0.0} & +0.7 & -3.7 & \textbf{0.0} & +0.7 & \textbf{0.0} & -- & \textbf{0.0} \\
Bushy\_Eyebrows & +1.0 & -0.7 & +0.7 & +3.7 & \textbf{0.0} & -0.3 & \textbf{0.0} & +1.0 & -0.7 & +0.7 & +3.7 & -- & -- & \textbf{0.0} \\
Chubby & \textbf{0.0} & \textbf{0.0} & \textbf{0.0} & +0.3 & \textbf{0.0} & +0.7 & \textbf{0.0} & \textbf{0.0} & \textbf{0.0} & \textbf{0.0} & +0.3 & \textbf{0.0} & \textbf{0.0} & \textbf{0.0} \\
Double\_Chin & \textbf{0.0} & \textbf{0.0} & \textbf{0.0} & \textbf{0.0} & \textbf{0.0} & \textbf{0.0} & \textbf{0.0} & \textbf{0.0} & \textbf{0.0} & \textbf{0.0} & \textbf{0.0} & \textbf{0.0} & \textbf{0.0} & \textbf{0.0} \\
Eyeglasses & +2.7 & \textbf{0.0} & \textbf{0.0} & +6.0 & \textbf{0.0} & \textbf{0.0} & \textbf{0.0} & +2.7 & \textbf{0.0} & \textbf{0.0} & +6.0 & \textbf{0.0} & -- & \textbf{0.0} \\
Goatee & \textbf{0.0} & \textbf{0.0} & \textbf{0.0} & -0.3 & \textbf{0.0} & \textbf{0.0} & \textbf{0.0} & \textbf{0.0} & \textbf{0.0} & \textbf{0.0} & -0.3 & \textbf{0.0} & -- & \textbf{0.0} \\
Gray\_Hair & \textbf{0.0} & \textbf{0.0} & \textbf{0.0} & +2.0 & \textbf{0.0} & \textbf{0.0} & \textbf{0.0} & \textbf{0.0} & \textbf{0.0} & \textbf{0.0} & +2.0 & \textbf{0.0} & \textbf{0.0} & \textbf{0.0} \\
Heavy\_Makeup & +25.0 & +26.0 & +26.0 & +23.7 & \textbf{0.0} & \textbf{0.0} & \textbf{0.0} & +25.0 & +26.0 & +26.0 & +23.7 & -- & +25.3 & \textbf{+13.0} \\
High\_Cheekbones & +21.7 & +18.7 & +21.3 & +14.7 & \textbf{0.0} & \textbf{0.0} & \textbf{0.0} & +21.7 & -- & +21.3 & \textbf{+14.7} & -- & +17.0 & -- \\
Male & +37.3 & +37.7 & +36.3 & +33.7 & \textbf{0.0} & \textbf{0.0} & \textbf{0.0} & +37.3 & +37.7 & +36.3 & \textbf{+33.0} & -- & +37.3 & +37.3 \\
Mouth\_Slightly\_Open & +13.7 & +11.7 & +14.0 & +5.3 & \textbf{0.0} & +0.7 & -0.3 & +13.7 & +11.7 & +14.0 & +5.3 & \textbf{0.0} & +12.7 & -0.7 \\
Mustache & \textbf{0.0} & \textbf{0.0} & \textbf{0.0} & \textbf{0.0} & \textbf{0.0} & \textbf{0.0} & \textbf{0.0} & \textbf{0.0} & \textbf{0.0} & \textbf{0.0} & \textbf{0.0} & \textbf{0.0} & -- & \textbf{0.0} \\
Narrow\_Eyes & \textbf{0.0} & \textbf{0.0} & \textbf{0.0} & \textbf{0.0} & -- & \textbf{0.0} & \textbf{0.0} & \textbf{0.0} & \textbf{0.0} & \textbf{0.0} & \textbf{0.0} & \textbf{0.0} & \textbf{0.0} & \textbf{0.0} \\
No\_Beard & +7.0 & +1.7 & +5.7 & +8.0 & \textbf{0.0} & \textbf{0.0} & \textbf{0.0} & +7.0 & +1.7 & +5.7 & +8.0 & \textbf{0.0} & +2.3 & \textbf{0.0} \\
Oval\_Face & +0.3 & \textbf{0.0} & \textbf{0.0} & \textbf{0.0} & \textbf{0.0} & -1.7 & \textbf{0.0} & +0.3 & \textbf{0.0} & \textbf{0.0} & \textbf{0.0} & \textbf{0.0} & -4.0 & \textbf{0.0} \\
Pale\_Skin & \textbf{0.0} & \textbf{0.0} & \textbf{0.0} & \textbf{0.0} & \textbf{0.0} & \textbf{0.0} & \textbf{0.0} & \textbf{0.0} & \textbf{0.0} & \textbf{0.0} & \textbf{0.0} & \textbf{0.0} & \textbf{0.0} & \textbf{0.0} \\
Pointy\_Nose & -5.0 & -7.3 & -0.3 & +2.0 & \textbf{0.0} & \textbf{0.0} & \textbf{0.0} & -5.0 & -7.3 & -0.3 & +2.0 & \textbf{0.0} & -2.0 & \textbf{0.0} \\
Receding\_Hairline & \textbf{0.0} & \textbf{0.0} & \textbf{0.0} & +0.3 & \textbf{0.0} & \textbf{0.0} & \textbf{0.0} & \textbf{0.0} & \textbf{0.0} & \textbf{0.0} & +0.3 & \textbf{0.0} & \textbf{0.0} & \textbf{0.0} \\
Rosy\_Cheeks & +0.3 & \textbf{0.0} & \textbf{0.0} & +1.3 & \textbf{0.0} & \textbf{0.0} & \textbf{0.0} & +0.3 & \textbf{0.0} & \textbf{0.0} & +1.3 & \textbf{0.0} & -- & \textbf{0.0} \\
Sideburns & +0.3 & \textbf{0.0} & \textbf{0.0} & +0.7 & \textbf{0.0} & \textbf{0.0} & \textbf{0.0} & +0.3 & \textbf{0.0} & \textbf{0.0} & +0.7 & \textbf{0.0} & -0.3 & \textbf{0.0} \\
Smiling & +30.3 & +26.7 & +22.3 & +16.0 & \textbf{0.0} & -0.3 & \textbf{0.0} & +30.3 & +26.7 & +22.3 & \textbf{+17.3} & -- & +25.3 & -- \\
Straight\_Hair & \textbf{0.0} & +0.3 & \textbf{0.0} & +2.3 & \textbf{0.0} & \textbf{0.0} & \textbf{0.0} & \textbf{0.0} & +0.3 & \textbf{0.0} & +2.3 & -- & +0.3 & \textbf{0.0} \\
Wavy\_Hair & +14.0 & +10.7 & +8.0 & +15.3 & \textbf{0.0} & \textbf{0.0} & \textbf{0.0} & +14.0 & +10.7 & +8.0 & +15.3 & \textbf{0.0} & -- & \textbf{0.0} \\
Wearing\_Earrings & +2.7 & \textbf{0.0} & +0.3 & +6.7 & \textbf{0.0} & \textbf{0.0} & \textbf{0.0} & +2.7 & \textbf{0.0} & +0.3 & +6.7 & \textbf{0.0} & \textbf{0.0} & \textbf{0.0} \\
Wearing\_Hat & \textbf{0.0} & \textbf{0.0} & \textbf{0.0} & +2.3 & \textbf{0.0} & \textbf{0.0} & \textbf{0.0} & \textbf{0.0} & \textbf{0.0} & \textbf{0.0} & +2.3 & \textbf{0.0} & \textbf{0.0} & \textbf{0.0} \\
Wearing\_Lipstick & +40.0 & +38.0 & +39.0 & +27.7 & -4.3 & \textbf{0.0} & -0.3 & +40.0 & +38.0 & +39.0 & +30.0 & -- & +38.3 & \textbf{+28.7} \\
Wearing\_Necklace & -1.0 & \textbf{0.0} & \textbf{0.0} & \textbf{0.0} & \textbf{0.0} & \textbf{0.0} & \textbf{0.0} & -1.0 & \textbf{0.0} & \textbf{0.0} & \textbf{0.0} & \textbf{0.0} & \textbf{0.0} & \textbf{0.0} \\
Wearing\_Necktie & +1.3 & \textbf{0.0} & \textbf{0.0} & +2.0 & \textbf{0.0} & \textbf{0.0} & \textbf{0.0} & +1.3 & \textbf{0.0} & \textbf{0.0} & +2.0 & \textbf{0.0} & \textbf{0.0} & \textbf{0.0} \\
Young & +11.0 & +9.3 & +11.0 & +12.3 & \textbf{0.0} & +8.7 & \textbf{0.0} & +11.0 & +9.3 & +11.0 & +12.3 & \textbf{0.0} & +8.7 & +0.3 \\
\midrule
Mean (pp) & +6.4 & +5.1 & +5.5 & +5.3 & -0.4 & +0.7 & 0.0 & +6.4 & +4.7 & +5.5 & +5.3 & 0.0 & +5.8 & +2.5 \\
\bottomrule
\end{tabular}%
}
\end{table}

\section{Hyperparameters}
\label{sec:appendix_hyperparameters}

The per-sample local-radius-capped $\lambda_i$ (Eq.~\ref{eq:lambda_closed_form}) eliminates the global $\lambda$ from the proposed method. We fix the remaining step-controlling hyperparameters across all 119 settings (39 NLP $\times$ 80 CelebA) with no per-setting tuning. Values and the appendix sections that justify them:

\begin{table}[H]
\centering
\small
\begin{tabular}{l l l}
\toprule
Symbol & Value & Role / justification \\
\midrule
$\varepsilon$           & $0.1$  & local-radius fraction (App.~\ref{sec:appendix_spectral_intuition}) \\
$h$                     & $512$  & concept-probe hidden dimension \\
probe training steps    & $800$  & per probe refit \\
probe-refit interval    & $8$    & every 8 outer iterations \\
$H$                     & $60$   & outer iterations \\
$k_{\min}$              & $8$    & TwoNN intrinsic-dimension floor \\
$\lambda_{\max}$        & $64$   & safety cap on per-row $\lambda_i$ (App.~\ref{sec:appendix_lambda_stats}) \\
CovMatch rank           & $2$    & number of covariance-asymmetry eigenvectors (App.~\ref{sec:appendix_preprocessing_derivation}) \\
$\alpha$                & $1$    & local-$\sigma$ exponent (App.~\ref{sec:appendix_alpha_regimes}) \\
\bottomrule
\end{tabular}
\end{table}

Every method in Tab.~\ref{tab:main_results_unified} runs at these fixed values.

\paragraph{Local-basis rank $r$ and the floor $k_{\min} = 8$.}
The rank $r$ of the local tangent basis is set from the intrinsic dimension of the clean representations $\mathbf{X}^{(0)}$, estimated once with TwoNN \citep{facco-etal-2017-two-nn} and kept fixed throughout the iterative loop:
\[
r = \max\bigl(k_{\min},\; \lceil \mathrm{TwoNN}(\mathbf{X}^{(0)}) \rceil\bigr), \qquad k_{\min} = 8.
\]
We use TwoNN as a data-driven default for the rank so that the basis is wide enough to capture the local variance of nearby representations but not so wide that it absorbs noise directions, both of which would degrade the spectrally-weighted direction in Eq.~\ref{eq:tangent_direction}. The floor of $8$ is a guard against pathological TwoNN outputs: on small or noisy representation pools the estimator can collapse to single-digit values (sometimes $1$--$3$), at which point a rank-$r$ basis is too short for the local singular spectrum to support a meaningful $\sigma$-weighting. The floor sets a conservative minimum that prevents this collapse without interfering with reasonable estimates; on our 119 settings TwoNN typically returns values well above $8$, so the floor does not bind in practice.

\section{Computational cost}
\label{sec:appendix_latency}

\Cref{tab:method_latency} reports the wall-clock cost of fitting and applying each method on a representative NLP panel (Qwen2.5-1.5B, gender, layer~14), using each method's canonical configuration from \Cref{tab:main_results_unified}. All nine methods are timed sequentially in a single job on one NVIDIA~B200 GPU with 8 CPU cores. We time only the fit-and-apply path: diagnostic probes used solely to log the $S$/$Y$ trajectories are excluded, while probes internal to a method's own stopping rule (INLP's convergence check, \MANCE{}'s scorer margin) are included.

\MANCE{} is the most expensive method, at roughly eight minutes per panel ($458.8$--$474.9$\,s across the three variants): each round re-estimates a local tangent basis per row (kNN against the fixed natural reference $\mathbf{X}^{(0)}$, then a batched SVD), and periodically refits the nonlinear scorer. The closed-form preprocessing in \MANCEp{} and \MANCEpp{} adds only a few seconds on top of the loop ($458.8 \to 470.5 \to 474.9$\,s). The baselines are far cheaper: LEACE and CovMatch are one-shot at a few seconds ($3.7$ and $7.3$\,s); the boundary and gradient erasers Obliviator, IGBP, and \AmbCEpp{} run in $20$--$24$\,s; and INLP takes $286.9$\,s, dominated by its CPU-bound per-round logistic-regression fits. This is a one-time, per-panel offline cost and does not affect inference: every method outputs an edited representation set, or an affine map for the closed-form erasers.

\begin{table}[t]
\centering
\caption{\textbf{Fit-and-apply wall-clock latency.} One representative panel (Qwen2.5-1.5B, gender, layer 14), every method under its canonical configuration from \Cref{tab:main_results_unified}, all timed sequentially on the same dedicated GPU (one NVIDIA B200, 8 CPU cores). Diagnostic per-round probes used only for reporting are excluded. INLP's fit is CPU-bound (per-round logistic-regression fits), so its wall-clock scales with the CPU allocation rather than the GPU.}
\label{tab:method_latency}
\small
\begin{tabular}{lr}
\toprule
Method & Wall-clock (s) \\
\midrule
LEACE & 3.7 \\
LEACE\,+\,CovMatch & 7.3 \\
\AmbCEpp{} & 19.8 \\
IGBP & 20.3 \\
Obliviator & 23.6 \\
INLP & 286.9 \\
\MANCE{} & 458.8 \\
\MANCEp{} & 470.5 \\
\MANCEpp{} & 474.9 \\
\bottomrule
\end{tabular}
\end{table}

\section{Why $\varepsilon = 0.1$ transfers across panels}
\label{sec:appendix_spectral_intuition}

Section~\ref{sec:closed_form_lambda} chooses the per-sample step size $\lambda_i$ so that the displacement magnitude satisfies $\|\tilde{\mathbf{x}}_i - \mathbf{x}_i\|_2 \le \varepsilon \cdot r_i$, where $r_i$ is the mean Euclidean distance from $\mathbf{x}_i$ to its $k$ nearest neighbors among the natural representations $\mathbf{X}^{(0)}$ (Eq.~\ref{eq:r_i_def}). This appendix explains why a single value $\varepsilon = 0.1$ can be used across all 39 NLP settings (13 language models $\times$ 3 concepts at 50\%-depth) and all 80 CelebA settings (40 attributes $\times$ 2 surgicality regimes) without per-setting tuning of the local-radius cap.

\paragraph{Reading the constraint.}
With $\|\tilde{\mathbf{x}}_i - \mathbf{x}_i\|_2 \le \varepsilon \cdot r_i$, the budget reads ``the displacement L2 norm is at most $\varepsilon$ fraction of the local Euclidean neighborhood radius.'' The local radius $r_i$ is recomputed every round: the neighbors are drawn from the fixed natural representations $\mathbf{X}^{(0)}$ but measured from the sample's current (edited) position $\mathbf{x}_i^{(t-1)}$, so as a point moves under earlier edits its distance to the surrounding natural neighborhood (and hence the budget) tracks the new location. The same $\varepsilon$ corresponds to a small move when the neighborhood is dense (small $r_i$) and a larger move when the neighborhood is sparse (large $r_i$), which is the behavior we want from a local first-order step.

\paragraph{Why $\varepsilon$ is setting-independent.}
A single global step size $\lambda$ has setting-specific units: applied verbatim to a Gemma-3-27B representation with $d=5376$ it moves the point by a much larger Euclidean distance than the same constant applied to a Qwen2.5-0.5B representation with $d=896$. The local $r_i$ used here is computed from each panel's own representations, so the displacement magnitude $\varepsilon \cdot r_i$ is automatically expressed in the panel's representation scale. The ratio of the displacement to the local neighborhood radius is the same across panels for a fixed $\varepsilon$, even though the absolute Euclidean distance moved is very different. This is the property that makes $\varepsilon = 0.1$ behave consistently across the 119 settings.

\paragraph{Per-row, not just per-panel.}
Because $r_i$ is computed at the level of each individual sample, the budget also adapts within a panel. Rows that lie in dense regions (where neighbors are close) get small allowed displacements; rows in sparse regions get larger ones. This is what distinguishes the per-row $\lambda_i$ in Eq.~\ref{eq:lambda_closed_form} from a global $\lambda$: a global step size has to be small enough not to over-edit the densest rows, which leaves the sparse rows under-edited. The per-row local-radius cap addresses both at once.

\section{CelebA control-set construction (least- vs.\ most-correlated five attributes)}
\label{sec:appendix_celeba_controls}

For each of the 40 binary CelebA attributes used as a target concept, we report two complementary surgicality regimes (Section~\ref{sec:celeba}, Tabs.~\ref{tab:celeba_per_panel_breakdown_dy1}--\ref{tab:celeba_per_panel_breakdown_dy5}, Figure~\ref{fig:per_attr_bars_celeba}): the \emph{5 least-correlated} attributes and the \emph{5 most-correlated} attributes, both selected from the remaining 39 attributes. This appendix records exactly how the two control sets are computed.

\paragraph{Computation.}
Let $\mathbf{Y} \in \{0, 1\}^{N_{\text{train}} \times 40}$ be the train-split label matrix, with $\mathbf{Y}_{i,k}$ the label of training example $i$ on attribute $k$. We compute the empirical Pearson correlation matrix
\begin{equation}
\mathbf{R}_{k\ell} \;=\; \frac{\mathrm{Cov}(\mathbf{Y}_{:,k}, \mathbf{Y}_{:,\ell})}{\sqrt{\mathrm{Var}(\mathbf{Y}_{:,k}) \cdot \mathrm{Var}(\mathbf{Y}_{:,\ell})}}
\quad \text{for } k, \ell \in \{1, \ldots, 40\}.
\end{equation}
For each target attribute $k$, we then rank the remaining $39$ attributes by absolute correlation $|\mathbf{R}_{k\ell}|$:
\begin{itemize}
\item the \emph{5 least-correlated} controls are the bottom 5 by $|\mathbf{R}_{k\ell}|$. They share \emph{minimal} information with the target and isolate clean surgicality.
\item the \emph{5 most-correlated} controls are the top 5 by $|\mathbf{R}_{k\ell}|$. They share \emph{substantial} information with the target and form a stricter test of whether erasure can preserve genuinely related downstream features.
\end{itemize}
The target attribute itself is excluded from the candidate pool (i.e., $\ell \neq k$). Pairs are computed in float64 with no shrinkage. 

\paragraph{Concrete examples.}
For \emph{Male}, the 5 most-correlated attributes are \emph{Wearing\_Lipstick} ($|r| \approx 0.78$), \emph{No\_Beard}, \emph{Heavy\_Makeup}, \emph{Attractive}, and \emph{5\_o\_Clock\_Shadow}; the 5 least-correlated are \emph{Narrow\_Eyes}, \emph{Blurry}, \emph{Straight\_Hair}, \emph{Receding\_Hairline}, and \emph{Pale\_Skin} (all $|r| < 0.05$). Erasing \emph{Male} while preserving the most-correlated controls is genuinely hard because the controls are partial proxies for the target; erasing while preserving the least-correlated controls tests whether the intervention is geographically narrow but not whether it disentangles correlated information. We therefore report both regimes side by side: each one isolates a different failure mode.

\paragraph{Full per-target listing.} Tabs.~\ref{tab:celeba_controls_least}--\ref{tab:celeba_controls_most} list the two control sets for every one of the 40 target attributes, each with its absolute Pearson $|r|$ on the train split. 

\begin{table}[t]
\centering
\caption[CelebA least-correlated control sets]{\textbf{CelebA least-correlated control sets (highly disentangled control concepts).} For each of the 40 target attributes (rows), the 5 control attributes selected as the bottom-5 by absolute Pearson $|r|$ on the train split, with each $|r|$ value in parentheses. These are the 5 attributes that share the least information with the target and isolate clean surgicality.}
\label{tab:celeba_controls_least}
\scriptsize
\setlength{\tabcolsep}{2.5pt}
\resizebox{\textwidth}{!}{%
\begin{tabular}{l|lllll}
\toprule
Target & \multicolumn{5}{c}{5 least-correlated controls (with $|r|$)} \\
\midrule
5\_o\_Clock\_Shadow & Receding\_Hairline (0.00) & Double\_Chin (0.01) & Young (0.01) & Bald (0.01) & Blurry (0.02) \\
Arched\_Eyebrows & Oval\_Face (0.00) & Bushy\_Eyebrows (0.01) & Bangs (0.01) & Receding\_Hairline (0.02) & Black\_Hair (0.03) \\
Attractive & Straight\_Hair (0.00) & Bushy\_Eyebrows (0.01) & Black\_Hair (0.01) & Mouth\_Slightly\_Open (0.04) & Wearing\_Necklace (0.08) \\
Bags\_Under\_Eyes & Straight\_Hair (0.00) & Wearing\_Necklace (0.01) & Big\_Lips (0.01) & Wearing\_Hat (0.01) & Black\_Hair (0.02) \\
Bald & Wearing\_Hat (0.00) & Mouth\_Slightly\_Open (0.01) & Smiling (0.01) & 5\_o\_Clock\_Shadow (0.01) & High\_Cheekbones (0.03) \\
Bangs & Black\_Hair (0.00) & Brown\_Hair (0.01) & Mouth\_Slightly\_Open (0.01) & Narrow\_Eyes (0.01) & Arched\_Eyebrows (0.01) \\
Big\_Lips & Chubby (0.00) & Mustache (0.00) & Goatee (0.00) & Bags\_Under\_Eyes (0.01) & Double\_Chin (0.01) \\
Big\_Nose & Wearing\_Necklace (0.01) & Straight\_Hair (0.03) & Rosy\_Cheeks (0.03) & Wearing\_Earrings (0.03) & Black\_Hair (0.05) \\
Black\_Hair & Bangs (0.00) & Smiling (0.00) & Mouth\_Slightly\_Open (0.01) & Attractive (0.01) & Chubby (0.01) \\
Blond\_Hair & Blurry (0.00) & Straight\_Hair (0.01) & Pale\_Skin (0.03) & Oval\_Face (0.04) & Narrow\_Eyes (0.05) \\
Blurry & Blond\_Hair (0.00) & Sideburns (0.00) & Goatee (0.00) & No\_Beard (0.01) & Wearing\_Hat (0.01) \\
Brown\_Hair & Wearing\_Necklace (0.00) & Mouth\_Slightly\_Open (0.00) & Bangs (0.01) & Straight\_Hair (0.02) & High\_Cheekbones (0.02) \\
Bushy\_Eyebrows & Chubby (0.00) & Narrow\_Eyes (0.00) & Smiling (0.00) & Wearing\_Hat (0.01) & Arched\_Eyebrows (0.01) \\
Chubby & Big\_Lips (0.00) & Bushy\_Eyebrows (0.00) & Mouth\_Slightly\_Open (0.01) & Wearing\_Necklace (0.01) & High\_Cheekbones (0.01) \\
Double\_Chin & 5\_o\_Clock\_Shadow (0.01) & Rosy\_Cheeks (0.01) & Wearing\_Necklace (0.01) & Big\_Lips (0.01) & Sideburns (0.02) \\
Eyeglasses & Straight\_Hair (0.01) & Blurry (0.01) & Black\_Hair (0.02) & Narrow\_Eyes (0.03) & Pale\_Skin (0.03) \\
Goatee & Blurry (0.00) & Big\_Lips (0.00) & Double\_Chin (0.02) & Receding\_Hairline (0.03) & Black\_Hair (0.03) \\
Gray\_Hair & Sideburns (0.01) & Narrow\_Eyes (0.02) & Bushy\_Eyebrows (0.02) & Smiling (0.02) & Straight\_Hair (0.03) \\
Heavy\_Makeup & Pale\_Skin (0.07) & Receding\_Hairline (0.07) & Straight\_Hair (0.07) & Narrow\_Eyes (0.07) & Brown\_Hair (0.10) \\
High\_Cheekbones & Chubby (0.01) & Straight\_Hair (0.02) & Brown\_Hair (0.02) & Young (0.02) & Bushy\_Eyebrows (0.02) \\
Male & Narrow\_Eyes (0.02) & Blurry (0.02) & Straight\_Hair (0.05) & Receding\_Hairline (0.08) & Pale\_Skin (0.10) \\
Mouth\_Slightly\_Open & Brown\_Hair (0.00) & Chubby (0.01) & Bangs (0.01) & Bald (0.01) & Black\_Hair (0.01) \\
Mustache & Big\_Lips (0.00) & Pale\_Skin (0.01) & Blurry (0.01) & Narrow\_Eyes (0.02) & Bald (0.03) \\
Narrow\_Eyes & Wearing\_Earrings (0.00) & Bushy\_Eyebrows (0.00) & Sideburns (0.00) & Bangs (0.01) & No\_Beard (0.01) \\
No\_Beard & Blurry (0.01) & Narrow\_Eyes (0.01) & Straight\_Hair (0.02) & Double\_Chin (0.03) & Receding\_Hairline (0.04) \\
Oval\_Face & Pointy\_Nose (0.00) & Straight\_Hair (0.00) & Arched\_Eyebrows (0.00) & Receding\_Hairline (0.01) & Wavy\_Hair (0.01) \\
Pale\_Skin & Wearing\_Hat (0.00) & Mustache (0.01) & Wearing\_Necktie (0.02) & Wavy\_Hair (0.02) & Bushy\_Eyebrows (0.02) \\
Pointy\_Nose & Oval\_Face (0.00) & Straight\_Hair (0.00) & Receding\_Hairline (0.01) & 5\_o\_Clock\_Shadow (0.03) & Pale\_Skin (0.03) \\
Receding\_Hairline & 5\_o\_Clock\_Shadow (0.00) & Rosy\_Cheeks (0.00) & Straight\_Hair (0.01) & Oval\_Face (0.01) & Pointy\_Nose (0.01) \\
Rosy\_Cheeks & Receding\_Hairline (0.00) & Double\_Chin (0.01) & Chubby (0.02) & Narrow\_Eyes (0.03) & Bushy\_Eyebrows (0.03) \\
Sideburns & Blurry (0.00) & Narrow\_Eyes (0.00) & Gray\_Hair (0.01) & Straight\_Hair (0.01) & Wearing\_Necktie (0.01) \\
Smiling & Bushy\_Eyebrows (0.00) & Black\_Hair (0.00) & Straight\_Hair (0.00) & Bald (0.01) & Brown\_Hair (0.02) \\
Straight\_Hair & Oval\_Face (0.00) & Bags\_Under\_Eyes (0.00) & Wearing\_Necklace (0.00) & Pointy\_Nose (0.00) & Smiling (0.00) \\
Wavy\_Hair & Oval\_Face (0.01) & Pale\_Skin (0.02) & Narrow\_Eyes (0.02) & Double\_Chin (0.02) & Sideburns (0.03) \\
Wearing\_Earrings & Narrow\_Eyes (0.00) & Receding\_Hairline (0.02) & Double\_Chin (0.02) & Chubby (0.03) & Oval\_Face (0.03) \\
Wearing\_Hat & Pale\_Skin (0.00) & Bald (0.00) & Bushy\_Eyebrows (0.01) & Wearing\_Necktie (0.01) & Bags\_Under\_Eyes (0.01) \\
Wearing\_Lipstick & Straight\_Hair (0.04) & Narrow\_Eyes (0.05) & Pale\_Skin (0.08) & Receding\_Hairline (0.09) & Black\_Hair (0.11) \\
Wearing\_Necklace & Brown\_Hair (0.00) & Straight\_Hair (0.00) & Bags\_Under\_Eyes (0.01) & Big\_Nose (0.01) & Double\_Chin (0.01) \\
Wearing\_Necktie & Wearing\_Hat (0.01) & Sideburns (0.01) & Pale\_Skin (0.02) & Blurry (0.02) & Smiling (0.03) \\
Young & 5\_o\_Clock\_Shadow (0.01) & Wearing\_Necklace (0.02) & Mouth\_Slightly\_Open (0.02) & High\_Cheekbones (0.02) & Wearing\_Hat (0.03) \\
\bottomrule
\end{tabular}%
}
\end{table}

\begin{table}[t]
\centering
\caption[CelebA most-correlated control sets]{\textbf{CelebA most-correlated control sets (highly entangled control concepts).} For each of the 40 target attributes (rows), the 5 control attributes selected as the top-5 by absolute Pearson $|r|$ on the train split, with each $|r|$ value in parentheses. These share substantial information with the target and form the surgicality stress test.}
\label{tab:celeba_controls_most}
\scriptsize
\setlength{\tabcolsep}{2.5pt}
\resizebox{\textwidth}{!}{%
\begin{tabular}{l|lllll}
\toprule
Target & \multicolumn{5}{c}{5 most-correlated controls (with $|r|$)} \\
\midrule
5\_o\_Clock\_Shadow & No\_Beard (0.53) & Male (0.42) & Wearing\_Lipstick (0.35) & Heavy\_Makeup (0.29) & Sideburns (0.27) \\
Arched\_Eyebrows & Wearing\_Lipstick (0.46) & Heavy\_Makeup (0.44) & Male (0.42) & Attractive (0.28) & Big\_Lips (0.28) \\
Attractive & Wearing\_Lipstick (0.51) & Heavy\_Makeup (0.50) & Male (0.45) & Young (0.42) & Arched\_Eyebrows (0.28) \\
Bags\_Under\_Eyes & Big\_Nose (0.35) & Male (0.30) & Heavy\_Makeup (0.28) & Wearing\_Lipstick (0.26) & Young (0.25) \\
Bald & Gray\_Hair (0.28) & Young (0.22) & Chubby (0.22) & Big\_Nose (0.22) & Wearing\_Necktie (0.19) \\
Bangs & Male (0.20) & Wearing\_Lipstick (0.20) & No\_Beard (0.15) & Heavy\_Makeup (0.14) & Receding\_Hairline (0.13) \\
Big\_Lips & Arched\_Eyebrows (0.28) & Male (0.21) & Wearing\_Lipstick (0.21) & Wavy\_Hair (0.18) & Wearing\_Necklace (0.17) \\
Big\_Nose & Male (0.36) & Bags\_Under\_Eyes (0.35) & Young (0.33) & Double\_Chin (0.31) & Chubby (0.29) \\
Black\_Hair & Bushy\_Eyebrows (0.29) & Brown\_Hair (0.24) & Blond\_Hair (0.24) & Young (0.15) & 5\_o\_Clock\_Shadow (0.14) \\
Blond\_Hair & Male (0.31) & Wearing\_Lipstick (0.31) & Heavy\_Makeup (0.28) & Black\_Hair (0.24) & Arched\_Eyebrows (0.20) \\
Blurry & Attractive (0.18) & Heavy\_Makeup (0.16) & Wearing\_Lipstick (0.15) & Young (0.11) & Arched\_Eyebrows (0.09) \\
Brown\_Hair & Black\_Hair (0.24) & Blond\_Hair (0.17) & Male (0.14) & Wearing\_Lipstick (0.13) & Wavy\_Hair (0.13) \\
Bushy\_Eyebrows & Black\_Hair (0.29) & No\_Beard (0.25) & Male (0.23) & 5\_o\_Clock\_Shadow (0.17) & Wearing\_Lipstick (0.16) \\
Chubby & Double\_Chin (0.46) & Young (0.30) & Big\_Nose (0.29) & Wearing\_Necktie (0.22) & Bald (0.22) \\
Double\_Chin & Chubby (0.46) & Young (0.33) & Big\_Nose (0.31) & Gray\_Hair (0.25) & Bags\_Under\_Eyes (0.24) \\
Eyeglasses & Wearing\_Lipstick (0.24) & Chubby (0.21) & Male (0.21) & Heavy\_Makeup (0.20) & Attractive (0.20) \\
Goatee & No\_Beard (0.56) & Sideburns (0.51) & Mustache (0.43) & Male (0.30) & Wearing\_Lipstick (0.26) \\
Gray\_Hair & Young (0.35) & Bald (0.28) & Double\_Chin (0.25) & Wearing\_Necktie (0.23) & Receding\_Hairline (0.23) \\
Heavy\_Makeup & Wearing\_Lipstick (0.82) & Male (0.68) & Attractive (0.50) & Arched\_Eyebrows (0.44) & Wearing\_Earrings (0.37) \\
High\_Cheekbones & Smiling (0.69) & Mouth\_Slightly\_Open (0.44) & Wearing\_Lipstick (0.29) & Wearing\_Earrings (0.29) & Male (0.27) \\
Male & Wearing\_Lipstick (0.80) & Heavy\_Makeup (0.68) & No\_Beard (0.52) & Attractive (0.45) & 5\_o\_Clock\_Shadow (0.42) \\
Mouth\_Slightly\_Open & Smiling (0.53) & High\_Cheekbones (0.44) & Wearing\_Earrings (0.22) & Rosy\_Cheeks (0.15) & Arched\_Eyebrows (0.13) \\
Mustache & Goatee (0.43) & No\_Beard (0.43) & Sideburns (0.36) & Male (0.24) & Big\_Nose (0.20) \\
Narrow\_Eyes & Big\_Lips (0.16) & Oval\_Face (0.14) & Big\_Nose (0.13) & Double\_Chin (0.12) & Mouth\_Slightly\_Open (0.09) \\
No\_Beard & Goatee (0.56) & 5\_o\_Clock\_Shadow (0.53) & Sideburns (0.52) & Male (0.52) & Wearing\_Lipstick (0.43) \\
Oval\_Face & Heavy\_Makeup (0.22) & High\_Cheekbones (0.21) & Attractive (0.21) & Smiling (0.20) & Rosy\_Cheeks (0.17) \\
Pale\_Skin & Attractive (0.12) & Male (0.10) & Young (0.09) & High\_Cheekbones (0.09) & Smiling (0.09) \\
Pointy\_Nose & Wearing\_Lipstick (0.25) & Heavy\_Makeup (0.25) & Male (0.24) & Arched\_Eyebrows (0.23) & Attractive (0.18) \\
Receding\_Hairline & Gray\_Hair (0.23) & Big\_Nose (0.22) & Young (0.18) & Double\_Chin (0.15) & Bags\_Under\_Eyes (0.13) \\
Rosy\_Cheeks & Heavy\_Makeup (0.28) & Arched\_Eyebrows (0.26) & Wearing\_Lipstick (0.25) & High\_Cheekbones (0.25) & Smiling (0.24) \\
Sideburns & No\_Beard (0.52) & Goatee (0.51) & Mustache (0.36) & Male (0.28) & 5\_o\_Clock\_Shadow (0.27) \\
Smiling & High\_Cheekbones (0.69) & Mouth\_Slightly\_Open (0.53) & Rosy\_Cheeks (0.24) & Wearing\_Earrings (0.23) & Wearing\_Lipstick (0.22) \\
Straight\_Hair & Wavy\_Hair (0.31) & Black\_Hair (0.14) & Wearing\_Hat (0.10) & Wearing\_Earrings (0.09) & Blurry (0.09) \\
Wavy\_Hair & Wearing\_Lipstick (0.35) & Heavy\_Makeup (0.33) & Male (0.33) & Straight\_Hair (0.31) & Attractive (0.24) \\
Wearing\_Earrings & Wearing\_Lipstick (0.37) & Heavy\_Makeup (0.37) & Male (0.37) & High\_Cheekbones (0.29) & Arched\_Eyebrows (0.26) \\
Wearing\_Hat & Wearing\_Lipstick (0.17) & Heavy\_Makeup (0.17) & Attractive (0.13) & Big\_Nose (0.12) & Male (0.12) \\
Wearing\_Lipstick & Heavy\_Makeup (0.82) & Male (0.80) & Attractive (0.51) & Arched\_Eyebrows (0.46) & No\_Beard (0.43) \\
Wearing\_Necklace & Male (0.26) & Wearing\_Lipstick (0.24) & Arched\_Eyebrows (0.23) & Wearing\_Earrings (0.17) & Big\_Lips (0.17) \\
Wearing\_Necktie & Male (0.30) & Wearing\_Lipstick (0.24) & Young (0.24) & Gray\_Hair (0.23) & Big\_Nose (0.22) \\
Young & Attractive (0.42) & Gray\_Hair (0.35) & Double\_Chin (0.33) & Big\_Nose (0.33) & Male (0.32) \\
\bottomrule
\end{tabular}%
}
\end{table}

\paragraph{Why two regimes per concept.}
Reporting only one control set hides whichever regime the method happens to be weak in. A method that destroys the entire representation will pass the least-correlated test (controls are essentially independent of the edit anyway) and fail the most-correlated test catastrophically; conversely, a method that performs no erasure will pass both. The difference between the two regimes per method is therefore informative even when both individual numbers look acceptable. Section~\ref{sec:celeba} reads the difference as a measure of how much an erasure method confuses ``avoid this concept'' with ``avoid everything correlated with it''.

\section{Closed-form per-row $\lambda_i$ statistics}
\label{sec:appendix_lambda_stats}

For each setting and each outer iteration of the proposed method (the manifold-constrained loop with fixed local-radius fraction $\varepsilon = 0.1$), the closed-form rule of Eq.~\ref{eq:lambda_closed_form} produces a per-row step magnitude $\lambda_i$ (capped at $\lambda_{\max} = 64$). Table~\ref{tab:lambda_stats} reports the distribution of these per-row $\lambda_i$ averaged across all 60 iterations of each trajectory and then aggregated across settings.

\begin{table}[t]
\centering
\small
\caption{\textbf{Per-row $\lambda_i$ statistics for the proposed method ($\varepsilon = 0.1$, $\lambda_{\max}=64$) on the 39-setting NLP grid at 50\%-depth.} Each cell averages the per-row $\lambda_i$ first across all 60 outer iterations of one trajectory, then across the 13 model settings of the concept. The within-setting $p_{10}/p_{90}$ spread (mean across settings) is roughly $14/45$, i.e.\ a $\sim$10$\times$ ratio between the least- and most-confident rows in a single iteration. The safety cap $\lambda_{\max}=64$ is observed only sporadically (max 57.32 across 39$\times$60 = 2340 iterations); the closed-form rule rarely binds against it.}
\label{tab:lambda_stats}
\begin{tabular}{l r r r r r}
\toprule
Concept & $n$ & mean $\lambda_i$ & median & min & max \\
\midrule
Sycophancy & 13 & 29.58 & 37.15 & 4.89 & 49.38 \\
Gender & 13 & 25.28 & 23.76 & 0.45 & 45.08 \\
Safety & 13 & 33.07 & 33.30 & 1.71 & 57.32 \\
\midrule
\textbf{Overall} & \textbf{39} & \textbf{29.31} & \textbf{30.19} & \textbf{0.45} & \textbf{57.32} \\
\bottomrule
\end{tabular}
\end{table}

The mean across the entire grid, $\bar{\lambda} \approx 29.31$, is the value at which \AmbCEpp{} appears in Tab.~\ref{tab:main_results_unified}: forcing the unconstrained full-space method to take the same effective step magnitude that the closed-form per-row rule produces under \MANCEpp{} isolates the gradient-direction constraint (tangent-projected vs.\ full-space) as the only remaining axis of variation between the two methods.

\section{Empirical support for the manifold premise}
\label{sec:appendix_geometry_contrast}

The geometric premise used by MCH (\Cref{sec:mch}) states that natural representations concentrate on a low-dimensional manifold. We check this directly, independent of any erasure result. Tab.~\ref{tab:geometry_contrast_summary} reports the local intrinsic dimension (TwoNN, \citealp{facco-etal-2017-two-nn}) and the mean angle between a single global concept direction and the local per-sample concept directions, for the primary nonlinear case (sycophancy) and the diagnostic linear baseline (gender). For sycophancy the intrinsic dimension is $34$--$53$, far below the dimension $d \in [768, 5376]$ of the Euclidean representation space, and the large local/global angles show that a single global direction is a poor first-order summary across neighborhoods, both consistent with concentration on a curved, low-dimensional structure rather than a global linear subspace.

\begin{table}[t]
\centering
\caption{Geometry contrast between sycophancy and gender across completed layers 4, 8, and 12. This is a focused contrast between the primary nonlinear case and the diagnostic baseline, not an all-dataset geometry census. The table reports two quantities: a local intrinsic-dimension range and the mean angle between a global concept direction and local per-sample concept directions. Under the manifold framing, large local-global angles indicate that a single global direction is a poor first-order summary across neighborhoods. They do not by themselves provide a direct estimate of curvature. Sycophancy shows substantially larger local-global misalignment than gender, supporting it as the primary case for locally adaptive, manifold-motivated erasure methods.}
\label{tab:geometry_contrast_summary}
\small
\begin{tabular}{lcc}
\toprule
Concept & Local intrinsic dimension & Mean local/global angle \\
\midrule
Sycophancy & 34--53 & 76.9--82.6$^\circ$ \\
Gender & 365--461 & 51.8--55.8$^\circ$ \\
\bottomrule
\end{tabular}
\end{table}

\section{Anisotropy regimes for the local-$\sigma$ spectrum}
\label{sec:appendix_alpha_regimes}

\paragraph{Where $\boldsymbol\sigma$ enters the update.}
The spectral exponent $\alpha$ controls how the singular values of the mean-centered local PCA matrix $\mathbf{S}_i$ enter the tangent direction $\mathbf{d}_i$ (Eq.~\ref{eq:tangent_direction}): the tangent coordinate of $\mathbf{u}_i$ along $\mathbf{v}_i^{(\ell)}$ is reweighted by $\sigma_{i,\ell}^{\alpha}$ before re-assembly in the Euclidean representation space. The local-radius budget itself (Eq.~\ref{eq:lambda_closed_form}) is direction-agnostic; it is a fraction of the local Euclidean neighborhood radius $r_i$. So $\alpha$ is what makes the update direction-aware in the local spectrum; the budget is what makes the update magnitude scale-aware in the local representation geometry.

Eq.~\ref{eq:tangent_direction} reweights each tangent coordinate $a_{i,k}$ by $\sigma_{i,k}^{\alpha}$, where $\alpha \in \mathbb{R}$ is an anisotropy exponent and $\sigma_{i,k}$ are the singular values of the mean-centered local PCA matrix $\mathbf{S}_i$ (Eq.~\ref{eq:local_pca_svd}). Three non-negative values of $\alpha$ have natural interpretations:
\begin{itemize}
\item \textbf{$\alpha = 0$}: the tangent direction $\mathbf{d}_i = \mathbf{B}_i \mathbf{c}_i$ is the bare projection of the unit gradient onto the local tangent basis; the singular spectrum plays no role in shaping the direction.
\item \textbf{$\alpha = 1$}: each tangent direction is reweighted by $\sigma_{i,k}$. \textbf{This is the default in this paper.} It uses the local PCA spectrum directly: high-$\sigma$ tangent axes receive more step mass, while thin axes receive less.
\item \textbf{$\alpha = 2$}: each tangent direction is reweighted by $\sigma_{i,k}^2$. This is a stronger SVD-spectrum weighting with a Mahalanobis-style intuition, but it is not derived by solving a covariance-constrained optimization: we do not form or invert a local covariance matrix. Geometrically, it suppresses low-variance tangent directions most strongly, which may also suppress residual concept information carried by thin directions.
\end{itemize}

\paragraph{Mahalanobis-style intuition (and what $\alpha = 1$ is \emph{not}).}
A reader familiar with Mahalanobis whitening might wonder whether our $\alpha = 1$ is the same operation. It is not. Classical Mahalanobis whitening explicitly uses a covariance inverse square root, $\mathbf{x}^{\mathrm{white}} = \mathbf{C}^{-1/2} \mathbf{x}$; in the local-PCA coordinates this corresponds to scaling each tangent direction by $\sigma_{i,k}^{-1}$, i.e.\ $\alpha = -1$ in our notation. Whitening makes every direction equal-weighted, which would dilute the concept-specific signal we want to suppress and force every tangent axis to receive equal step magnitude.

We use a \emph{positive} power, $\sigma_{i,k}^{+1}$, which does the opposite: it reweights the tangent step \emph{toward} high-$\sigma$ principal axes (the directions where the local manifold has the most support and where the linear approximation is most reliable) and away from low-$\sigma$ thin tail directions where the linear approximation is less reliable and motion most easily leaves the manifold. The three non-negative values $\alpha \in \{0, 1, 2\}$ form a one-parameter family interpolating between (i) treating every tangent direction as equally reliable ($\alpha = 0$, isotropic), (ii) reweighting by $\sigma_{i,k}$ to respect the local ellipsoid mildly ($\alpha = 1$, our default), and (iii) a stronger spectral weighting that follows the local ellipsoid more aggressively and suppresses thin directions most ($\alpha = 2$). Whitening ($\alpha = -1$) sits outside this range for our problem because its sign is opposite the geometric correction we want; we therefore do not include it.

\newpage

\end{document}